\pdfoutput=1
\PassOptionsToPackage{hyphens}{url}
\documentclass[11pt]{article}
\usepackage[margin=1in]{geometry}
\usepackage{amsmath,amssymb,graphicx,booktabs,natbib,microtype,tabularx,float,array}
\usepackage{flafter}
\usepackage{placeins}
\usepackage[hidelinks]{hyperref}
\usepackage{caption}
\newcommand{\Hrow}{H^{W}_{\mathrm{row}}}
\newcommand{\Hcol}{H^{W}_{\mathrm{col}}}
\newcommand{\HW}{H^{W}}
\newcommand{\kraw}{k_{\mathrm{raw}}}
\newcommand{\kbi}{k_{\mathrm{bi}}}
\newcommand{\krow}{k_{\mathrm{row}}}
\newcommand{\kcol}{k_{\mathrm{col}}}
\newcommand{\kident}{k_{\mathrm{ident}}}
\newcommand{\dF}{\Delta_{F}}
\newcommand{\dkmix}{\Delta_{k}^{\mathrm{mix}}}

\title{A Mesoscopic View of Transformer Weights\\Through Row and Column Scale Fields}
\author{Tiexin Ding\thanks{tiexinding@gmail.com}}
\date{}

\begin{document}
\maketitle

\begin{abstract}
Pooled statistics of Transformer weights obscure how magnitude is distributed across functional
channels, while individual weights are too numerous to compare directly. We study the mesoscopic level
between them: row and column scale fields, the median-centred log-RMS profiles of a weight matrix over its
channels, which together with a global scale and a full balanced core represent the matrix exactly. Across
public Pythia checkpoints at four sizes and controlled runs from three initialization families, balancing
reveals similar measured core magnitude profiles. A mixture bridge, with its form fixed before the analysis
and its coefficients fitted, predicts the pooled-shape departure from field width on held-out runs and data
arms of the controlled grid. The indexed fields retain further structure: they align across projections
that share a functional channel, and query/key profiles follow reassigned RoPE frequencies rather than fixed
matrix coordinates. Training trajectories show early field formation followed by component-dependent
broadening or recession. Extending the channel-based analysis to AdamW's second moment reveals related
functional organization in its log-space row and column factors. Finally, edits of a frozen checkpoint
separate reciprocal scale balance, which preserves the forward computation, from relative channel gain:
flattening the gain increases in-distribution loss while preserving matrix norms and the balanced core.
Row and column scale fields thus connect pooled magnitude statistics to channel organization and provide
coordinates for tracking and testing trained weight structure.

\end{abstract}

\section{Introduction}

Transformer weights can be inspected entry by entry or summarized at the matrix level, but these
views answer different questions. Norms, spectral statistics and fitted magnitude distributions
support comparison across matrices, yet do not directly identify which functional channels carry
scale variation \citep{martin2021implicit,ding2026a_weibull}. In particular, pooling magnitudes
can mix variation within channels with differences in scale across channels. We study the
intermediate level at which those differences can be measured, aligned and followed through training.

\begin{figure}[H]
\centering
\includegraphics[width=\textwidth]{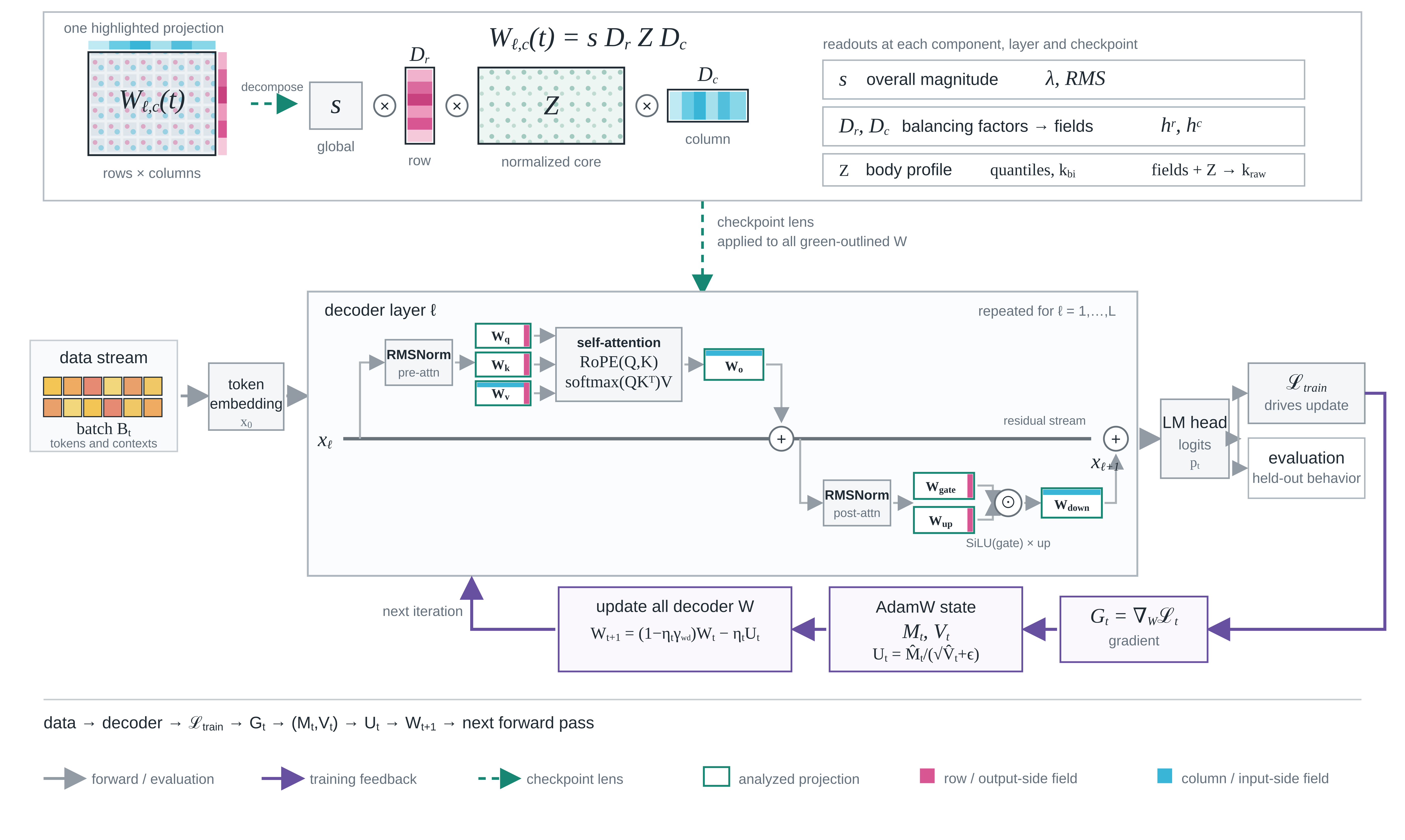}
\caption{\textbf{A mesoscopic view of Transformer weights through row and column scale fields.}
The training loop produces a sequence of projection matrices. At selected checkpoints, each matrix
is represented by its global RMS, balancing factors and full balanced core. Direct row and column
log-RMS fields are linked to these factors by the conversions in Section~\ref{sec:decomp}.
Field widths, core magnitude profiles and fitted pooled parameters are statistical read-outs of
this representation; the fitted scale $\lambda$ tracks the global RMS $s$ through the measured ratio
$\lambda/s$. Coloured strips mark the empirically dominant field sides; an unmarked side
need not be homogeneous. The dashed analysis path is not part of the training update.}
\label{fig:overview}
\end{figure}

Our objects are the row and column scale fields: the median-centred log-RMS profiles of a weight
matrix over its output and input channels. We relate these direct measurements to the balanced
representation $W=sD_rZD_c$, where $s=\mathrm{RMS}(W)$ and every row and column of $Z$ has unit RMS.
The complete representations are interconvertible when the full signed core is retained
(Section~\ref{sec:decomp}). For statistical analysis, the fields record channel-scale heterogeneity
and arrangement, while the magnitude profile of $Z$ describes what remains after balancing.
The fitted Weibull parameters serve as pooled read-outs rather than assumptions about every entry.
Figure~\ref{fig:overview} locates these objects in the decoder and its training loop.

This framework asks what remains after channel balancing, how the fields account for pooled
statistics and functional organization, and which changes in the fields affect the computation.
We address these questions with controlled LLaMA-style \citep{touvron2023llama} 70M runs, initialization-family
and RoPE-reassignment experiments, Pythia size comparisons \citep{biderman2023pythia}, optimizer-state records and frozen-checkpoint
edits. Their roles and replication levels differ and are specified in Table~\ref{tab:configs}.

\paragraph{Contributions.}
\begin{enumerate}\itemsep2pt
\item We connect direct channel-scale measurements to balanced-core analysis. Across the tested
settings, balancing reveals similar measured core magnitude profiles. A calibrated mixture bridge
predicts pooled-shape departure from field width on held-out runs and data arms within the controlled
grid, while retaining the distinction between complete fields and their scalar summaries.

\item We identify functional and temporal organization in the fields. Profiles match across
projections sharing a channel space, and a paired RoPE reassignment moves $q/k$ profiles with their
assigned frequencies. Training trajectories reveal broad early formation followed by component-dependent
continuation or recession, which terminal pooled statistics alone do not show.

\item We extend the channel-based analysis to AdamW's second moment and test the functional role
of paired weight fields. Log-space optimizer factors align with functional channel spaces.
Frozen-checkpoint edits separate an invariant reciprocal balance from a loss-sensitive relative
gain: flattening the gain increases loss at fixed matrix norms and balanced core, and exceeds the
tested Gaussian controls after displacement normalization.
\end{enumerate}

\section{Related work}
\label{sec:related}

\paragraph{Weight magnitude, direction, and channel scales.}
Weight normalization separates the length and the direction of a weight vector
\citep{salimans2016weightnorm}, and later work relates weight norms to effective learning rates,
decoupled weight decay, and rotational equilibrium, down to the single neuron
\citep{vanlaarhoven2017l2, loshchilov2019decoupled, wan2021spherical, kosson2024rotational}. Recent
language-model studies make channel magnitudes explicit optimization variables:
\citet{velikanov2026multipliers} attach learnable per-row and per-column multipliers to free the
weight-decay--noise equilibrium norm and report that they broaden the distribution of row norms in the
attention-input and gate projections, and \citet{haegele2026md} factorize each weight into a fixed-norm direction and learnable
per-row and per-column gains updated at separate rates. These methods make channel magnitudes explicit
training variables. We instead measure the row and column scale structure that emerges in ordinarily trained
checkpoints and relate it to pooled statistics, channel identities and controlled edits.

\paragraph{Weight-side and channel-level diagnostics.}
Trained weights have been read as a record of training through spectral statistics
\citep{martin2021implicit} and through the training-data mixture that can be recovered from fine-tuned
weights \citep{huang2026warp}. At the channel level, \citet{chen2024matthew} use the variance of channel
weight norms to assess layer width and to identify stages of training. Our analysis represents that
heterogeneity by complete row and column scale fields, together with the core obtained by two-sided RMS
balancing.

\paragraph{Row--column optimizer geometry.}
Row and column structure also appears in optimizer design, from Adafactor's row--column
reconstruction of the second moment \citep{shazeer2018adafactor} to the axis-wise preconditioners of
K-FAC and Shampoo \citep{martens2015kfac, gupta2018shampoo}, which are different decompositions; coordinatewise
normalization also brings Adam's update close to a smoothed sign direction \citep{balles2018dissecting,
kunstner2023noise}. Our analysis of AdamW's second moment is a descriptive fit to the existing state, asking
whether its row and column factors line up with the channel spaces found in trained weights.

\paragraph{Transformer channel identities and shared paths.}
The architecture gives many rows and columns an identifiable role. RoPE assigns a fixed frequency to
each coordinate pair of the query and key projections \citep{su2024roformer}, and these frequencies
are used non-uniformly across heads and dependency ranges \citep{barbero2024round, hong2024token}, while
context-extension methods rescale them according to wavelength \citep{peng2024yarn}; recent work relates
frequency usage to the training data \citep{wu2026ropedata} and makes the rotation frequencies themselves
learnable \citep{karypis2026lerope}.
Matrix-level analyses of the paired projections study the products rather than the individual factors:
\citet{saponati2025selfattention} find that autoregressive training induces high-norm columns in the
query--key product (in their token-row convention), and \citet{kobayashi2024lowrank} show that at stationary
points of the weight-decay-regularized loss the two factors of a bilinear product have equal Gram matrices,
which makes the penalty equivalent to a nuclear-norm penalty and lowers the rank of the query--key and
value--output products, and find this balance approximately satisfied, row by row, in pretrained Llama~2. Building on these observations, we ask
how the row and column scale fields of each individual projection follow architecture-defined
channel identities, how fields pair across matrices that share a channel space, and how a controlled
reassignment of RoPE frequencies moves the learned scale profile.

\paragraph{Relation to our earlier studies.}
Our earlier work introduced a protocol-matched Weibull description of pooled weight magnitudes,
decomposed AdamW scale growth, and related the resulting scale coordinate to pre-training data
statistics \citep{ding2026a_weibull, ding2026b_adamw, ding2026c_data}. The effort-versus-quality
boundary discussed in Section~\ref{sec:limitations} is from a companion study
that is under review, referred to below as the companion study. The present study moves from pooled
matrix coordinates to the intermediate row--column structure of the weights, using the pooled shape as
one macroscopic read-out while treating the scale fields and the core as the primary objects.

\section{Method: Row--Column Scale Fields}
\label{sec:framework}

Table~\ref{tab:notation} in Appendix~\ref{app:notation} lists the notation used throughout the paper, with the
equation or section that defines each symbol.

\subsection{Weight-matrix representations and conversions}
\label{sec:decomp}

We use three complete representations of each weight matrix $W \in \mathbb{R}^{d_{\mathrm{out}} \times d_{\mathrm{in}}}$,
\begin{equation}
W \;\longleftrightarrow\; (s,\,D_r,\,Z,\,D_c) \;\longleftrightarrow\; (s,\,h^r,\,h^c,\,Z),
\label{eq:chain}
\end{equation}
built from the following quantities.

\textbf{Global scale, balancing factors and core.} The exact re-parameterization
\begin{equation}
W = s\,D_r\,Z\,D_c, \qquad s = \mathrm{RMS}(W),
\label{eq:decomp}
\end{equation}
separates the global scale $s$, the positive diagonal balancing factors $D_r$ and $D_c$, and the balanced core $Z$,
\begin{equation}
\mathrm{RMS}_j(Z_{ij}) = 1 \;\text{ for every row } i, \qquad \mathrm{RMS}_i(Z_{ij}) = 1 \;\text{ for every column } j ,
\label{eq:balanced_core}
\end{equation}
obtained by alternating row and column RMS balancing of the Sinkhorn--Knopp type
\citep{sinkhorn1967, sinkhorn1967prescribed}.

\textbf{Scale fields.} With the channel RMS $R_i = \mathrm{RMS}_j(W_{ij})$ and $C_j = \mathrm{RMS}_i(W_{ij})$, the row and
column scale fields are their logarithms centred by the median over channels (the centring changes neither
the widths nor the profile correlations),
\begin{equation}
h^r_i = \log R_i - \operatorname{median}_i \log R_i, \qquad
h^c_j = \log C_j - \operatorname{median}_j \log C_j .
\label{eq:fields}
\end{equation}
The field width $\HW$ is the standard deviation of the field on a kind's identity side, marked in
Figure~\ref{fig:overview}.

\textbf{Conversions.} Given $s$, the fields determine the channel RMS: since the mean of $R_i^2$ over rows
and the mean of $C_j^2$ over columns both equal $s^2$,
\begin{equation}
R_i = \frac{s\,e^{h^r_i}}{\big(d_{\mathrm{out}}^{-1}\textstyle\sum_{i'} e^{2h^r_{i'}}\big)^{1/2}}, \qquad
C_j = \frac{s\,e^{h^c_j}}{\big(d_{\mathrm{in}}^{-1}\textstyle\sum_{j'} e^{2h^c_{j'}}\big)^{1/2}} .
\label{eq:fieldsinv}
\end{equation}
The fields and the balancing factors describe the same channel scales but are not entrywise equal: each
marginal RMS also depends on the $Z^2$-weighted squared factors of the opposite side. With the core retained
as a full signed matrix, positive diagonal scalings of $Z$ that meet these row and column RMS targets recover
the balancing factors, up to their reciprocal constant, and hence $W$; Appendix~\ref{app:conversions} gives
the coupling between factors and fields, the iteration and its uniqueness. These conversions hold for the
complete representations; the field widths and pooled shapes used below are summaries of them, and the
relation between those summaries is tested by the mixture bridge.

\paragraph{Global scale and fitted scale.}
Let $\lambda_{\mathcal P}$ and $k_{\mathcal P}$ denote the scale and shape returned by the fitting protocol
$\mathcal P$ (Appendix~\ref{app:fitting}). The ideal probability-plot fit is scale equivariant,
\begin{equation}
\lambda_{\mathcal P}(W) = s\,\lambda_{\mathcal P}(W/s), \qquad k_{\mathcal P}(W) = k_{\mathcal P}(W/s),
\label{eq:lambdaequiv}
\end{equation}
and with the fixed histogram grid used here the relation holds approximately: refitting $W/s$ reproduces
$\lambda/s$ within 0.2\% and $k$ within 0.004 (Appendix~\ref{app:lambdaequiv}). Every shape read-out
therefore concerns $W/s$, and $\lambda$ is $s$ times the dimensionless ratio $\lambda/s$.
For an exact Weibull magnitude distribution $s^2 = \lambda^2\,\Gamma(1+2/k)$, so
\begin{equation}
\lambda/s = \Gamma(1+2/k)^{-1/2},
\label{eq:weibullratio}
\end{equation}
which varies by only 4\% between $k = 1.17$ and $1.25$ (the protocol ratio itself is measured,
Appendix~\ref{app:lambdaequiv}). The fitted scale is therefore a read-out of the global scale: the growth
of $\lambda$ under AdamW and its dependence on the training data studied earlier
\citep{ding2026b_adamw, ding2026c_data} are growth of $s$, linked to $\lambda$ there through
eq.~\eqref{eq:weibullratio} at locked shape; eq.~\eqref{eq:lambdaequiv} does not assume the Weibull form.

\subsection{Mixture bridge}
\label{sec:bridgedef}

On a kind's identity side the field is the centred log-scale of the one-sided decomposition, shown here
for rows,
\begin{equation}
W = \mathrm{diag}(R)\,Z^r, \qquad \mathrm{RMS}_j(Z^r_{ij}) = 1 \;\text{ for every row } i,
\label{eq:onesided}
\end{equation}
with $R_i = \mathrm{RMS}_j(W_{ij})$. For row-identity kinds, $\kident = \krow$ is the fitted shape of $Z^r$; for
column-identity kinds, $\kident = \kcol$ is obtained from the corresponding column-normalized core. Under
approximate independence of the identity-axis field and its one-sided core the log-variances of the two add (Appendix~\ref{app:weibullref}),
which motivates the one-axis relation, with its form fixed before the analysis,
\begin{equation}
\dkmix \equiv \kraw^{-2} - \kident^{-2} \approx \alpha_P (\HW)^2 .
\label{eq:bridge}
\end{equation}
The exact-Weibull value $6/\pi^2 \approx 0.608$ (Appendix~\ref{app:weibullref}) serves as a reference, and the
protocol coefficient $\alpha_P$ is estimated empirically. We also test the two-axis extension
\begin{equation}
\kraw^{-2} - \kbi^{-2} \approx \alpha_r (\Hrow)^2 + \alpha_c (\Hcol)^2 ,
\label{eq:bridge2}
\end{equation}
an empirical relation, since each direct marginal also carries the opposite factors
(eq.~\eqref{eq:factorsfields}). Both models are fitted through the origin and evaluated by
leave-one-run-out and leave-one-data-arm-out prediction, in which the field width and the normalized-core
shape of each held-out matrix are measured inputs, so the bridge is a conditional prediction of the pooled
shape; identifiability checks are in
Appendix~\ref{app:corebridge}.

\subsection{Optimizer-state row--column factors}
\label{sec:vfactordef}

Section~\ref{sec:optimizer} reads the same two axes off AdamW's bias-corrected second moment $\hat V$
(Appendix~\ref{app:vfactors}, eq.~\eqref{eq:adamw_update}), taken from the dumped state at a checkpoint.
We fit
\begin{equation}
\log \hat V_{ij} = \mu + a_i + b_j + \varepsilon_{ij}, \qquad \textstyle\sum_i a_i = \sum_j b_j = 0,
\label{eq:vfactors}
\end{equation}
whose least-squares factors are the centred row and column means of $\log \hat V$, with $\varepsilon$ the
coordinate-specific remainder. We report the variance explained by the joint model and by each factor, and
correlate factors that index the same channel space within a layer (median over layers). The
factorization is descriptive; the update rule, the motivation of the model and the fits are in
Appendix~\ref{app:vfactors}.

\subsection{Evidence design}
\label{sec:sources}

Table~\ref{tab:configs} in Appendix~\ref{app:configs} lists each evidence source with its protocol and
its role. The evidence falls into three classes: controlled multi-seed comparisons, the paired RoPE
reassignment with three seed pairs, and single-seed state, boundary and edit analyses, which are descriptive.
The statistical unit is the training run; matrix fits within a run are not independent replications.

\section{Results}
\label{sec:results}

The results show how row and column scale fields connect individual weights to pooled matrix
statistics. After two-sided balancing, the tested components approach a common magnitude-profile
core; their pooled differences arise mainly from how scale is distributed over channels. The fields
follow architecture-defined identities, evolve differently across components, and have a
corresponding topology in AdamW's second-moment state. They therefore provide a mesoscopic
description of trained Transformer weights.

\subsection{Two-sided balancing exposes a common core}
\label{sec:core}

\begin{figure}[tbp]
\centering
\includegraphics[width=\textwidth]{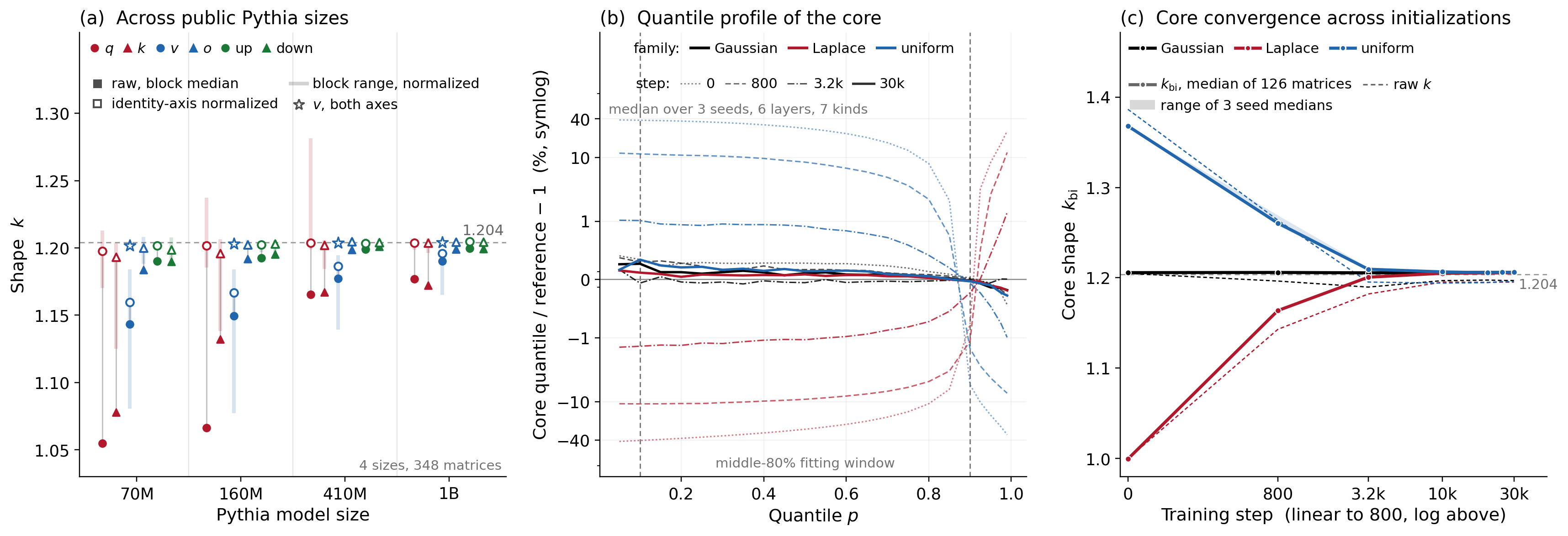}
\caption{\textbf{Two-sided balancing reveals a common core across components, model sizes and
initialization families.}
\textbf{(a)} Final public Pythia checkpoints; grey connectors pair the raw and identity-axis-normalized
fits of the same component kind (filled and open markers), stars mark two-sided balancing of $v$, and
coloured bars span the normalized block medians.
\textbf{(b)} Each quantile of the unit-RMS core $|Z|$ divided by the corresponding quantile of the
unit-RMS half-normal (the magnitude of a Gaussian weight), minus one, for one initialization family at one
checkpoint. At 30{,}000 steps each family lies within 0.26\% of it throughout the displayed range (median
pooled over the three families: 0.19\%).
\textbf{(c)} Two-sided core shape in nine LLaMA-style 70M runs. The horizontal line in (a) and (c) is
the reference shape of a Gaussian weight under the middle-80\% protocol, $k \approx 1.20$, derived for a
half-normal magnitude in \citet{ding2026a_weibull}.}
\label{fig:core}
\end{figure}

What remains after the row and column scale fields are removed? At the public Pythia endpoints,
the pooled component shapes differ, most visibly for $q$ and $k$. Normalizing each matrix along its
architecture-defined identity axis places the block medians of $q$, $k$, $o$ and both FFN projections
within 1.193--1.205 across all four model sizes. The $v$ projection is the exception under
one-sided normalization and enters the band only after both axes are balanced
(Figure~\ref{fig:core}(a)); this exception anticipates the two-sided channel organization of $v$
examined in Section~\ref{sec:identity}. The controlled grid gives the corresponding trajectory
result: after two-sided balancing, the central-body shape of all seven component kinds remains in
component medians $\kbi = 1.203$--$1.212$ at every checkpoint, while individual matrices spread wider
(Appendix~\ref{app:commoncore}; per-kind values in the data package). Most of the component separation visible in the pooled
distributions is therefore carried by the channel-scale fields rather than by the balanced core.

The full core profiles give a stronger view of this convergence. Gaussian, Laplace and uniform
initializations begin with clearly different magnitude profiles, but their two-sided cores approach
the same protocol-matched reference across the fitted body and through the displayed 0.99 quantile
(Figure~\ref{fig:core}(b)). The reference is the magnitude profile of a Gaussian weight, on which the
Gaussian family starts, so the substantive evidence comes from the Laplace and uniform families: they start about 40\% away in opposite directions and
converge to the same profile. Training therefore brings distinct initial core distributions into a
common measured body shape.

Figure~\ref{fig:core}(c) shows how this state is reached. The Gaussian core stays near the reference,
while the Laplace and uniform cores move progressively toward it and join it within the first
10{,}000 steps. This motion also shows that the common core is not imposed algebraically by the
balancing procedure: the balanced cores begin differently and converge during training. Scale-field
formation proceeds at the same time. By step 800 every identity-axis field is already several times its
initialization width (3.1--4.2 times in the Gaussian runs), while the non-Gaussian cores are still merging
(Figure~\ref{fig:timing}, middle and bottom rows, Appendix~\ref{app:dynamics}). Core convergence and field formation are
therefore concurrent rather than successive. Sections~\ref{sec:bridge}--\ref{sec:evolution} follow
these fields as they shape the pooled statistics, organize over channels and separate into distinct
training trajectories.

\subsection{Scale-field width predicts pooled-shape departure}
\label{sec:bridge}

Section~\ref{sec:core} read the representation in reverse: removing the row and column scale fields
exposed a common balanced core. Here we read it forward. Pooling channels that carry similar local
core profiles at different scales moves the pooled shape away from that core, and the mixture bridge
of Section~\ref{sec:bridgedef} quantifies this map back to the observed pooled statistic: the one-axis
form starts from the one-sided core shape $\kident$ of eq.~\eqref{eq:onesided}, the two-axis form from
the balanced core shape $\kbi$.

\begin{figure}[t]
\centering
\includegraphics[width=0.8\textwidth]{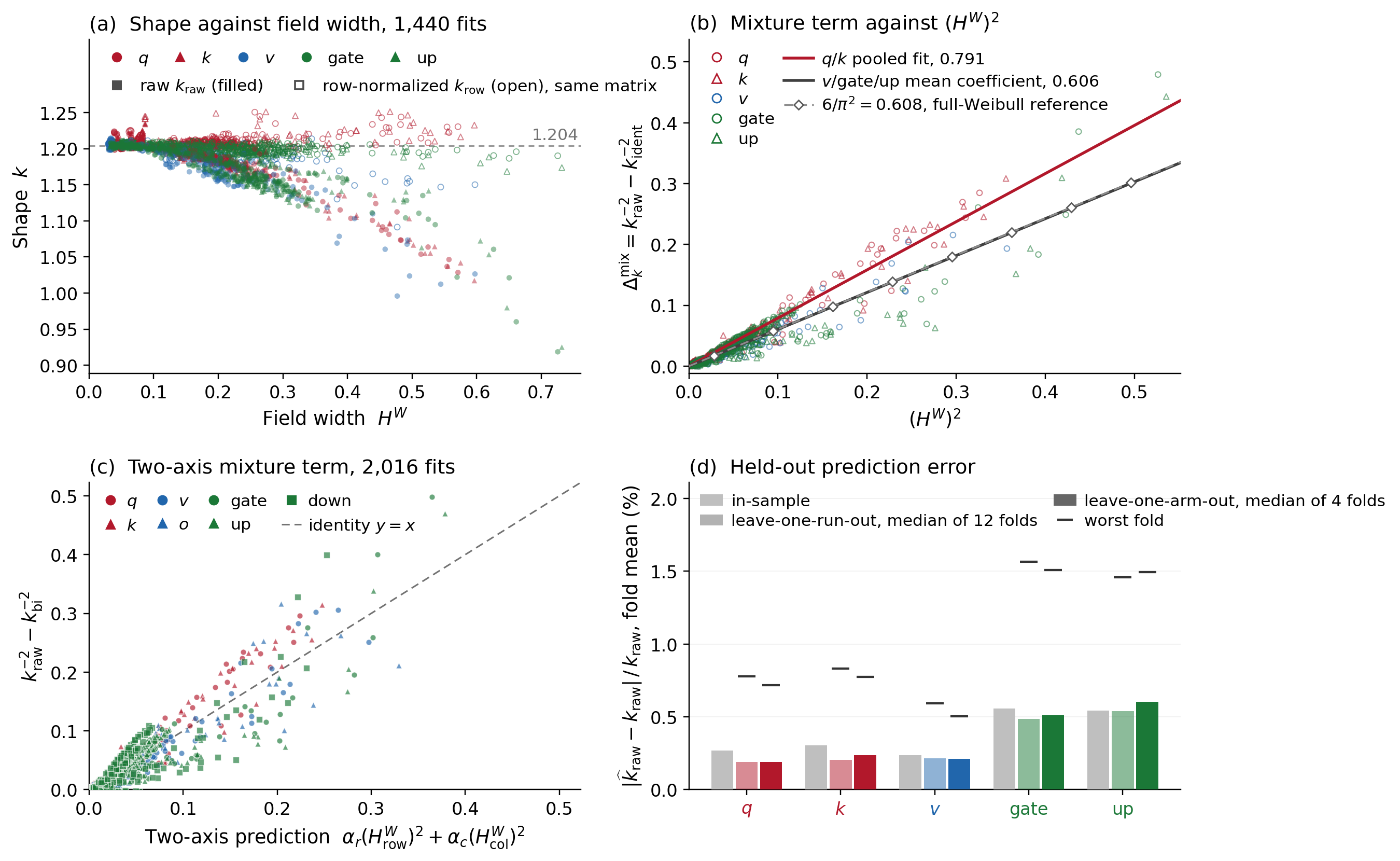}
\caption{\textbf{Scale-field mixing predicts the pooled-shape departure on the controlled grid.}
Panels (a), (b) and (d) use the 1{,}440 row-identity fits (4 data arms $\times$ 3 seeds $\times$ 4
checkpoints $\times$ 5 kinds $\times$ 6 layers) and panel (c) the two-axis fits of all seven kinds on the same runs;
colours and markers identify component kinds throughout. Panel (c) uses one pooled two-axis fit over the seven kinds. Panel (d) scores the one-axis prediction of
$\kraw$ with per-kind coefficients: bars give the in-sample error and the medians over held-out runs and
held-out arms of the fold-mean relative error, and marks show the worst fold; the field width and
normalized-core shape of each held-out matrix are measured inputs. Absolute leave-one-run-out errors of the
one- and two-axis models are in Figure~\ref{fig:twoaxis}(c).}
\label{fig:bridge}
\end{figure}

The effect is visible directly in Figure~\ref{fig:bridge}(a): as the row-field width grows, $\kraw$
departs from the reference while $\krow$ stays close to the core value. On the controlled grid the
relation~\eqref{eq:bridge} gives $\alpha_P = 0.791$ with $R_0^2 = 0.976$ for pooled
$q/k$ (Figure~\ref{fig:bridge}(b)), whereas $v$, gate and up sit at 0.59--0.64 with mean 0.606, close to
the exact-Weibull value $6/\pi^2 = 0.608$; the $q/k$ excess is associated with a scale--shape pairing term
(Appendix~\ref{app:corebridge}). The same direction holds in all 12 runs for each of the five row-side kinds, with per-kind
coefficients in Figure~\ref{fig:twoaxis}; the column-side kinds $o$ and down enter through the
two-axis form below. Refitted with one run or one whole
data arm held out, the bridge predicts the held-out $\kraw$ with a median error of 0.2--0.6\% by kind; the
worst held-out fold reaches about 1.6\%, in the shuffled high-repetition arm D3 (Figure~\ref{fig:bridge}(d)).
The relation is therefore not confined to the fitted trajectories.

Relative to the two-sided core, adding both field widths extends the bridge to all seven component
kinds (Figure~\ref{fig:bridge}(c)): the row-plus-column model~\eqref{eq:bridge2} gives
$R_0^2 = 0.904$ against 0.803 for the row-only model, with the gain concentrated in $v$, $o$ and
down, which connects the two-sided restoration of $v$ in Figure~\ref{fig:core}(a) to its measured
row and column fields. This extends the bridge to the column-identity kinds, with pooled coefficients of 0.67 for rows and 0.46 for columns. The bridge
retains the width of the field but not the arrangement of scales over channels
(Appendix~\ref{app:corebridge}); that information is examined in Section~\ref{sec:identity}.

\subsection{Architecture assigns the coordinates of scale fields}
\label{sec:identity}

Figure~\ref{fig:overview} distinguishes two levels of structural organization. The computation
graph determines whether a functional identity is indexed by matrix rows or columns, and
architecture-specific coordinates such as RoPE determine how that identity is arranged within the
selected axis. We test both levels through normalization, cross-projection profile matching, and a
paired RoPE reassignment.

\subsubsection{Functional identities map to matrix sides}
\label{sec:identity_sides}

Figure~\ref{fig:overview} gives the structural map for the scale fields. For a projection $y = Wx$,
rows index output channels and columns index input channels, and the gradient contributed by each token
has the outer-product form $\delta x^{\top}$ (Appendix~\ref{app:vfactors}). A functional identity therefore appears on the row side
of the matrix that produces it and on the column side of the matrix that consumes it
(Table~\ref{tab:paths}; the row-side and column-side fields are the two colours drawn on each
projection in Figure~\ref{fig:overview}).

\begin{table}[H]
\centering\footnotesize
\setlength{\tabcolsep}{4pt}
\begin{tabularx}{\textwidth}{@{}l X l >{\raggedright\arraybackslash}p{4.6cm}@{}}
\toprule
projection & functional identity & matrix side & normalization removing most of the departure \\
\midrule
$q$, $k$ & logit coordinate, RoPE frequency & rows & row \\
$v$ & head/value output; input-side channel profile & rows and columns & two-sided \\
$o$ & head/value input & columns & column \\
gate, up & FFN hidden output & rows & row \\
down & FFN hidden input & columns & column \\
$q$, $k$, $v$ & residual-stream input channel & columns & none (column fields widest in $v$; column normalization leaves $\kraw$ nearly unchanged) \\
\bottomrule
\end{tabularx}
\caption{Where the computation graph places each functional identity (matrix side, a structural fact), and
which one-sided normalization removes most of the pooled-shape departure on the controlled grid (an empirical
result; values in Table~\ref{tab:sides}).}
\label{tab:paths}
\end{table}

Consistent with this map, row normalization removes most of the pooled-shape departure of $q$, $k$, gate
and up and explains it with $R_0^2$ of 0.89--0.98, whereas column normalization does so for $o$ and down
(per-kind values in Table~\ref{tab:sides}, Appendix~\ref{app:corebridge}). The value projection has the two
fields closest in width (0.13 and 0.11 on the grid) and, in the public models, is the only kind that one-sided
normalization leaves clearly below the common core; two-sided balancing restores it. Its row field follows head/value identities, whereas its column
field carries a persistent input-side channel profile that is also visible in other read-side
projections. Why these two fields acquire comparable strength remains unresolved.

\subsubsection{Shared computational paths carry matched fields}
\label{sec:identity_paths}

\begin{figure}[tbp]
\centering
\includegraphics[width=\textwidth]{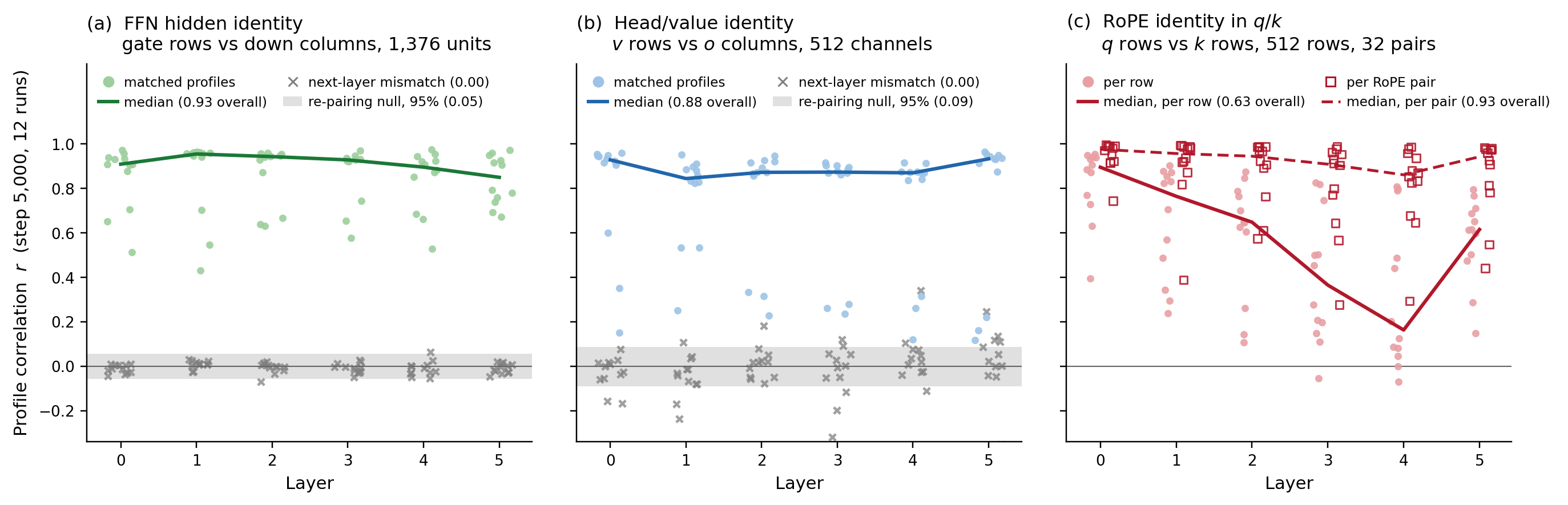}
\caption{\textbf{Shared functional channels carry matched scale-field profiles across projections.}
Correlations are computed at step 5{,}000 over the 12 controlled LLaMA-style 70M runs, per layer;
pale markers are single run--layer observations, lines connect per-layer medians over runs, and the overall
median over the 72 layer $\times$ run cells is given in each legend. The next-layer
mismatch pairs the same projections across adjacent layers. Up against down matches as closely as gate
against down (median $r$ 0.93). In (c), direct $q/k$ row matching is weaker and layer dependent, whereas
aggregating rows by RoPE frequency pair (32 pairs per head, median over heads) gives a consistently matched
profile. Long-run and data-arm
contrasts are reported in Appendix~\ref{app:axes}.}
\label{fig:transfer}
\end{figure}

Axis assignment also makes shared channels comparable: when two projections share a functional channel,
one profile can be read from the producing matrix's rows and the other from the consuming matrix's columns,
and whether training makes them match is an empirical question. Figure~\ref{fig:overview} contains three
such paths, $v \rightarrow o$, gate/up $\rightarrow$ down and $q \leftrightarrow k$, the last read on the
rows of both matrices and matched by RoPE pair, and Figure~\ref{fig:transfer} tests them.

All three pairings are observed: gate/up rows match down columns, $v$ rows match $o$
columns, and $q$ and $k$ rows match once indexed by RoPE pair. Layer-mismatched and within-layer
re-paired controls sit at the null, no pairing is present at initialization, and the data arms change
the strength of the matching but not its location. The gate--down match of step 5{,}000 weakens by
30{,}000 steps, when the persistent FFN pairing is up--down (Appendix~\ref{app:axes}). The fields follow the
shared functional channels.

\subsubsection{\texorpdfstring{RoPE indexes and couples the $q/k$ row-scale fields}{RoPE indexes and couples the q/k row-scale fields}}
\label{sec:identity_rope}

\begin{figure}[tbp]
\centering
\includegraphics[width=\textwidth]{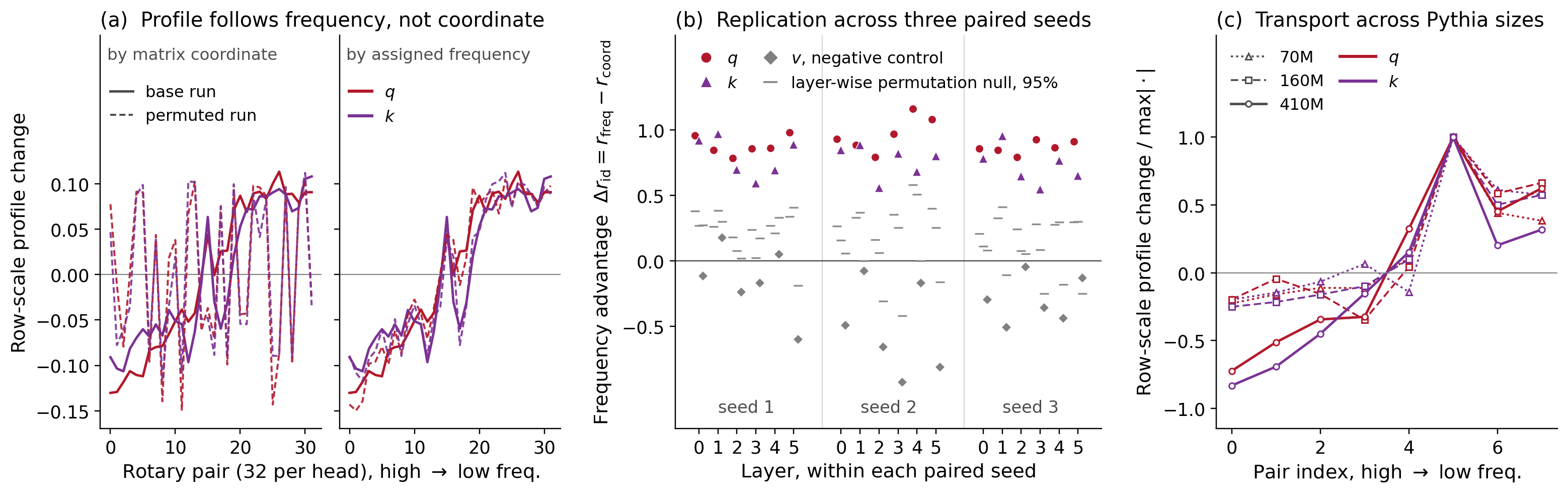}
\caption{\textbf{RoPE frequency indexes the $q/k$ row-scale profiles.} Profiles are cumulative
median-centred changes in log row RMS. \textbf{(a)} Base and frequency-permuted LLaMA-style 70M runs of the
representative seed pair, aggregated over heads and layers. \textbf{(b)} Frequency-indexing advantage
$\Delta r_{\mathrm{id}} = r_{\mathrm{freq}} - r_{\mathrm{coord}}$, where $r_{\mathrm{coord}}$ compares the paired
profiles at the same matrix coordinates and $r_{\mathrm{freq}}$ at the same assigned frequencies; one point per
layer and seed pair, with the $v$ control and the per-layer one-sided 95th-percentile permutation null.
\textbf{(c)} A separately trained Pythia-protocol size series (70M, 160M, 410M; one data arm, 8{,}000 steps;
rotary on 16 of 64 head dimensions, 8 pairs per head); each profile is divided by its maximum absolute value,
comparing shape and peak location rather than amplitude.}
\label{fig:rope}
\end{figure}

The dominant $q/k$ scale field lies on the row axis. We tested how this field is indexed by permuting
the assignment of RoPE frequencies while preserving the frequency set, architecture, initialization,
data order and seed (Figure~\ref{fig:rope}). In the representative pair, the base and permuted
profiles correlate at 0.98 for both $q$ and $k$ after frequency re-indexing, compared with 0.01 and
0.06 at the original matrix coordinates. Across three paired seeds, $\Delta r_{\mathrm{id}}$ is
positive and above the permutation null in all six layers for both projections. The $v$ control has no
reproducible positive frequency advantage (0.03, $-0.86$, $-0.15$ across seeds). RoPE frequency
therefore indexes the $q/k$ row-scale profiles.

The rotary construction explains why frequency, and not the matrix coordinate, indexes these fields.
Within a head, RoPE writes the logit between positions $m$ and $n$ as a sum over the rotary pairs $p$
of the query and key activations,
\[
\sum_p |q_p|\,|k_p| \cos\!\big((n-m)\theta_p + \phi_p\big),
\]
so $\theta_p$ sets how each pair's contribution varies with relative distance, and the gradient that
reaches the pair's weight rows carries the same rotation. A frequency thus assigns a computational role;
reassigning it to other rows moves the role, and the learned profile moves with it. The same construction
sets the unit on which a scale is defined. For distinct frequencies, the logits are unchanged by a common
rotation of the $q$ and $k$ rows of a pair, which mixes the two rows, and by the reciprocal rescaling
$q_p \to a\,q_p$, $k_p \to k_p/a$, the balance mode of Section~\ref{sec:gainedit}; without the rotation,
any invertible map $q \to Aq$, $k \to A^{-\top}k$ within a head leaves them unchanged. RoPE thus reduces
this freedom to one scale and one phase per pair: the scale of a single row changes under the rotation,
that of the pair does not, and the pair's $q$--$k$ gain is fixed up to the balance. This accounts for
pair aggregation matching $q$ and $k$ where single rows do not (Figure~\ref{fig:transfer}(c); the log-mean pair
read-out used there is not itself rotation invariant but agrees with the invariant pooled-RMS read-out,
Appendix~\ref{app:axes}) and for
the head-organized profile without RoPE. The symmetry fixes the unit, not the sign or strength of the
coupling: across pairs, the covariance of the $q$ and $k$ pair profiles is $[\mathrm{var}(g) - \mathrm{var}(b)]/4$, so positive
matching means that training varied the gain more than the balance, a measured outcome (Appendix~\ref{app:gainedit}), and $v/o$, with the same
freedom within a head, also couples (Figure~\ref{fig:transfer}(b)).

The frequency organization also appears across a separately trained Pythia-protocol size series. The
$q/k$ profiles correlate at 0.88--0.96 between 70M, 160M and 410M, and all six profiles reach their maximum
at pair index 5 (of 0--7).
Positive profile values denote frequencies whose rows grew more than the median row. Both the
LLaMA-style intervention and the Pythia series place the stronger relative growth toward the
lower-frequency part of their respective rotary grids.

The permutation separates field arrangement from its scalar summaries. Between paired runs, $\kraw$
and $\krow$ agree to within 0.5\%, while $\HW$ changes by different amounts across seeds
and layers (Figure~\ref{fig:rope_readout}). The full profile $h^r$ retains the frequency-to-row
assignment; $\HW$ summarizes its spread, and the pooled shape $k$ summarizes the resulting marginal shape.

The no-positional-encoding arm separates row-field formation from frequency organization. In this
run, $q$ and $k$ still develop substantial row-scale heterogeneity on a similar early timescale, but
the profile becomes primarily head-organized. The pair-level $q/k$ correlation falls from 0.90 to
0.39, still above its re-pairing null (97.5\% quantile 0.17 for this six-layer median), while the row-level
correlation rises from 0.26 to 0.85: without rotation each row coordinate is itself a unit of the $q \cdot k$
product. The $v/o$ and up/down paths remain well above their re-pairing nulls (Figure~\ref{fig:nope}).
Removing the rotary coordinate thus removes the frequency index and weakens the pair-level coupling, while
substantial row-scale heterogeneity remains. The comparison rests on one seed
and reaches a higher loss, so its magnitude differences remain descriptive.

Architecture supplies the channel coordinates of the scale fields. Section~\ref{sec:evolution} next
follows how the profiles written on those coordinates accumulate or recede during training.

\subsection{Scale fields form early in training but are retained selectively}
\label{sec:evolution}

\begin{figure}[tb]
\centering
\includegraphics[width=\textwidth]{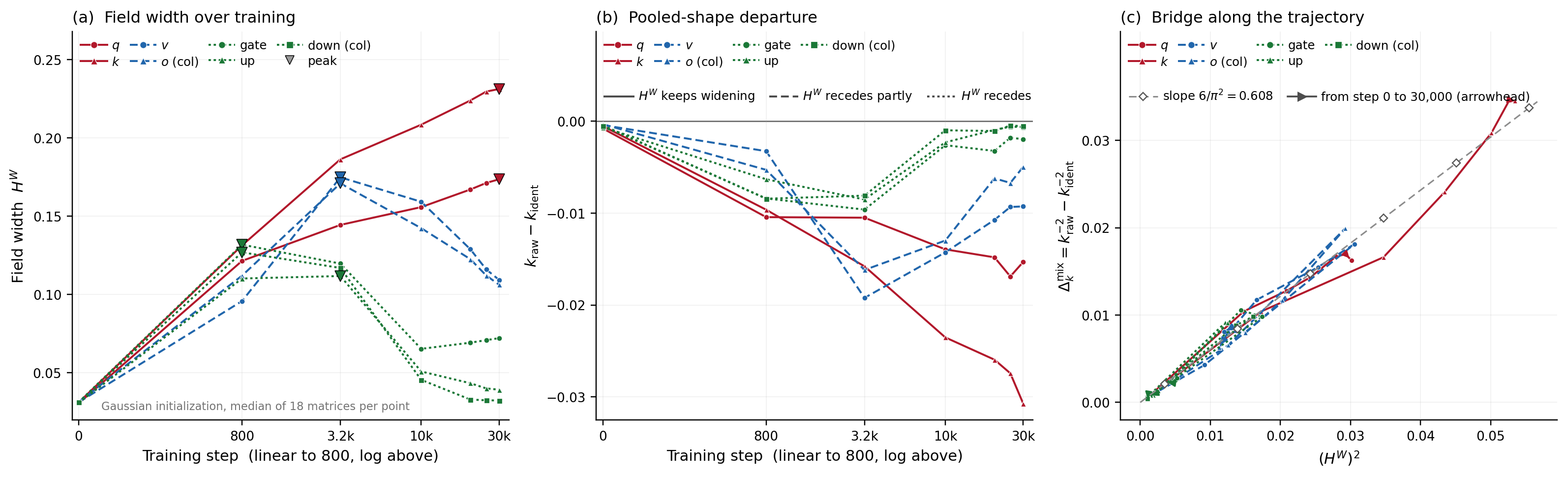}
\caption{\textbf{Scale fields form early in every projection kind, but their later retention differs.}
Each kind is read on its identity side, and $\kident$ is the fit after normalizing that side only; the $v$
curve therefore reports its row-side contribution, its two-sided restoration being given in
Section~\ref{sec:core}. Each point is the median over 18 matrices (six layers $\times$ three seeds of the
Gaussian family) at seven sampled steps; a matrix enters only if both fits pass $R^2 \geq 0.99$.
\textbf{(a)} Identity-axis field width; the common initial value is the finite-sample floor when each marginal
RMS is estimated from 512 entries, and triangles mark each kind's largest sampled width. \textbf{(b)} Shape
difference $\kraw - \kident$. \textbf{(c)} Mixture term $\kraw^{-2} - \kident^{-2}$ against $(\HW)^2$; in these
median trajectories the receding kinds return along nearly the same path.}
\label{fig:evolution}
\end{figure}

We follow the identity-axis field widths from initialization to step 30{,}000 in the three
Gaussian-initialized runs, sampled at steps 0, 800, 3{,}200, 10{,}000, 20{,}000, 25{,}000 and 30{,}000;
each point is the median over six layers and three seeds (Figure~\ref{fig:evolution}). The fields are not
built monotonically toward their terminal form. The sampled trajectories fall into three phases, and a
terminal checkpoint records only the last.
\textbf{Formation} (steps 0--800): by the first sampled checkpoint every projection kind has left the
initialization floor, at $\HW \approx 0.10$--$0.13$, while the balanced cores of the three initialization
families are still converging (Section~\ref{sec:core}).
\textbf{Separation} (800--3{,}200): the fields diverge by kind; $q$ and $k$ keep widening, $v$ and $o$ reach
their largest sampled width at step 3{,}200, and the FFN fields stop growing, gate and down peaking
already at step 800 and up nearly flat.
\textbf{Selective retention} (3{,}200--30{,}000): $q$ and $k$ widen further, to 0.174 and 0.231; $v$ and $o$
narrow to about 62\% of their peak and are still narrowing at step 30{,}000; the FFN fields narrow
substantially, gate with a small late rebound and down close to its initialization width (per-kind values
in Appendix~\ref{app:dynamics}). Formation is common and retention is selective: the terminal differences
between kinds reflect continued widening, partial recession or a return toward the initialization width,
not whether a field formed. The same qualitative sequence appears in all three initialization families.
Retention and recession here describe the width of the identity-axis field, not the preservation of
individual channel profiles or a return of weights to their initial values.

Across the sampled trajectories, the pooled shape tracks the contemporaneous field width. Along each
kind's median trajectory, $\dkmix$ against $(\HW)^2$ is linear through the origin with slopes 0.59--0.73
and $R_0^2 \geq 0.98$, and the widening and receding branches show little separation (Figure~\ref{fig:evolution}(c)); the slopes sit
below the across-matrix coefficient 0.791 of the grid, so the coefficient depends on what is compared.
As the FFN fields recede, the departure of the raw from the identity-axis fit closes to within 0.002,
inside the spread of the reference, while the retained $q/k$ fields keep it at $-0.015$ and $-0.031$
(Figure~\ref{fig:evolution}(b)). A near-reference terminal shape therefore does not mean that a projection
stayed homogeneous: the FFN fields broadened early and later narrowed.

The later trajectories also depend on the training condition. Without positional encoding
(Figure~\ref{fig:nope}) the $q$ field still widens, to 0.203 against 0.171, and becomes head-organized;
from step 3{,}200 to 30{,}000 $v$ holds its width (0.185 to 0.186) and $o$ declines only modestly (0.210 to
0.192), whereas the FFN fields recede again (Appendix~\ref{app:axes}). This run uses one seed and reaches a different
loss, so it marks a boundary condition, not a cause. FFN recession is the most reproducible trajectory
across the tested conditions, and what retains the attention-side fields remains open.

\FloatBarrier
\subsection{Extending channel-scale analysis to AdamW's second moment}
\label{sec:optimizer}

\begin{figure}[tbp]
\centering
\includegraphics[width=0.85\textwidth]{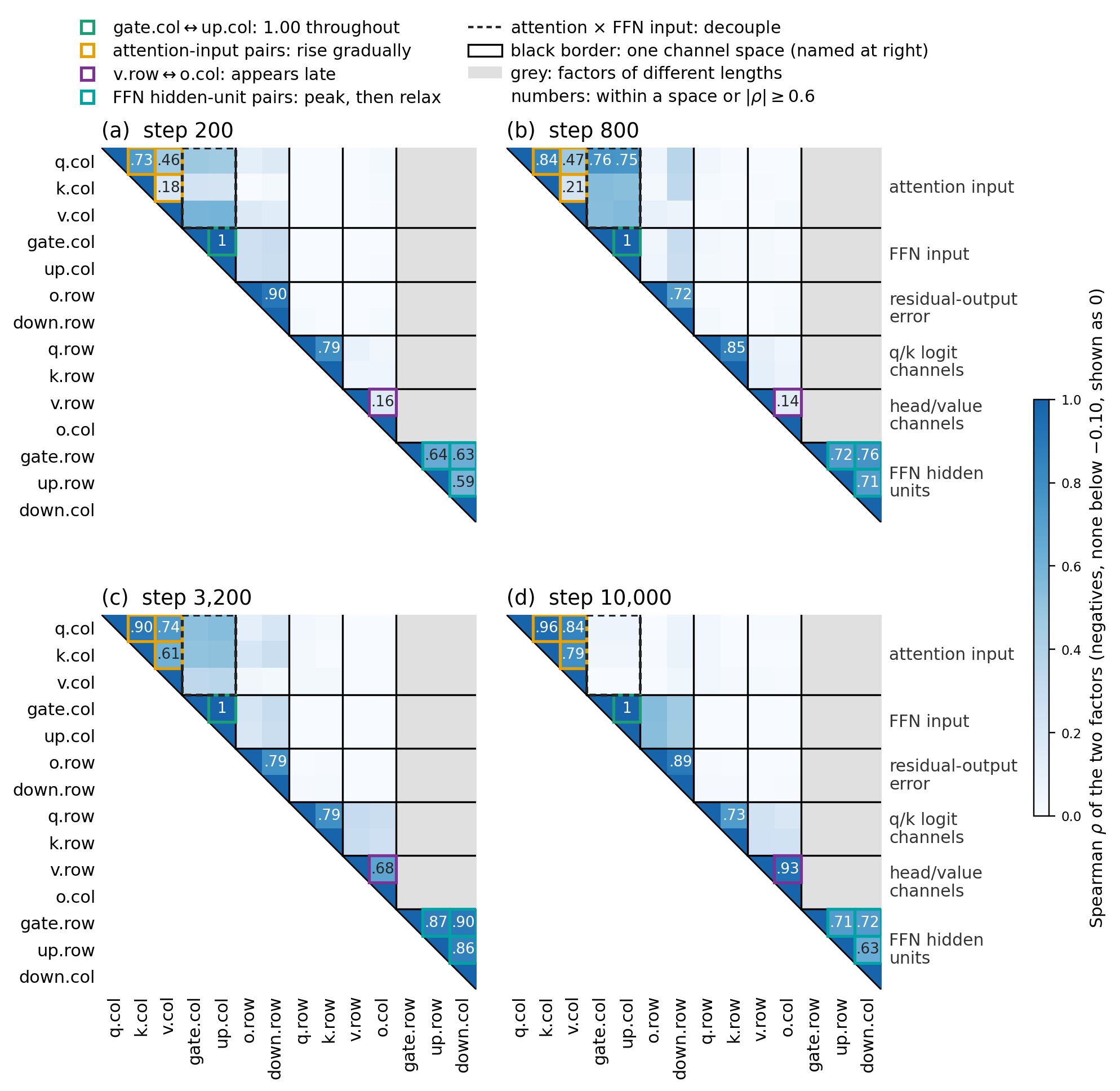}
\caption{\textbf{The row and column factors of AdamW's second moment align with the decoder's functional
channel spaces.} Spearman correlation $\rho$ among the 14 row and column factors of $\log \hat V$ for the
seven projection kinds at steps 200, 800, 3{,}200 and 10{,}000 of one LLaMA-style 70M run (single seed; the
step-400 dump enters the 210 fits but is not drawn); each factor is the centred mean of $\log \hat V$ along
its axis, and each cell the median over six layers. Grey cells pair factors of different length and are
undefined; negative values, none below $-0.10$, are drawn as 0. At step 10{,}000 the median correlation is
$+0.81$ over the 10 within-space pairs and $+0.01$ over the 48 comparable across-space pairs; the permutation
null (0.07) is the median over cells of the 95th percentile of $|\rho|$ under 200 within-layer permutations.}
\label{fig:topology}
\end{figure}

The row and column axes of the analysis are not specific to the weights. Every per-parameter array of a
projection, whether gradient, first moment or second moment, is indexed by the same output and input
channels, so the channel-scale read-out carries over; only the statistic has to suit the array. For
AdamW's bias-corrected second moment $\hat V$, the positive state from which the adaptive denominator is built, the
factors $a_i, b_j$ of Equation~\eqref{eq:vfactors} are the row and column main effects of $\log \hat V$,
not the log-RMS fields $h$ of the weights, but they index the same channels. We fit them to one
LLaMA-style 70M run (single seed) at five steps, 210 matrix--step fits (seven kinds $\times$ six layers
$\times$ five steps). The model accounts for a median 0.92 of the variance of $\log \hat V$, with a
minimum of 0.76, and the split between the two factors follows the weight-side identity axis: the row factor
dominates for $q$ and $k$ and, less sharply, for gate and up, the column factor for $o$ and down, and the two
are comparable for $v$ (Appendix~\ref{app:vfactors}).

The factors align with the functional channel spaces. Figure~\ref{fig:topology} groups the fourteen row and
column factors by the architecture-defined functional channel space each one indexes, a grouping fixed
before any correlation is computed. At step 10{,}000 the median correlation is $+0.81$ between factors of the same channel space and
$+0.01$ between factors of different spaces, against a permutation null of 0.07. The block structure is
pronounced but not complete: the columns that read the FFN input and the rows that carry the residual-output
error become correlated during training, whereas the attention-input columns do not (Appendix~\ref{app:vfactors}).

The blocks form on different time courses, marked in Figure~\ref{fig:topology}: the pair that reads one
tensor, the gate and up columns, is already fully correlated at the first dump, the
attention-input and value-path pairs strengthen over training, and the FFN hidden-unit block peaks at step
3{,}200 and partly relaxes, the direction of the FFN weight fields in Section~\ref{sec:evolution}. Factors
attached to the same nominal residual coordinate but read at different normalized tensors, the attention
input and the FFN input, correlate strongly at step 800 and hardly at all by step 10{,}000: the factors come to
track the tensor at a computational position, not the coordinate index. The optimizer state therefore carries the fixed connectivity of the architecture together
with structure that accumulates path by path during training.

The channel-scale analysis thus extends beyond the weights: on the optimizer state the same axes show
structure aligned with the functional channel spaces. The correspondence describes the adaptive
denominator, not a generator of the weight fields (Section~\ref{sec:limitations}). We next test on a frozen
checkpoint which changes in the paired scale fields preserve the computation.

\FloatBarrier
\subsection{Paired scale-field edits separate functional gain from invariant balance}
\label{sec:gainedit}

For the paired identity-axis profiles $h_1$ and $h_2$ of a shared channel, the gain $g=h_1+h_2$ is the
log of the relative product of the two channel scales and the balance $b=h_1-h_2$ the log of their
ratio. On the controlled grid at step 5{,}000, $\mathrm{var}(g)/\mathrm{var}(b)$ is 14--25 on the matched paths,
against about 1 at initialization and for re-paired $v/o$ and up/down profiles (the $q/k$ re-pairing null is
wider, 0.5--2.0; Appendix~\ref{app:gainedit}): the shared variation is concentrated in the gain. Each edit
rewrites all six layers of one path, $q/k$, $v/o$ or up/down, in the 30{,}000-step base checkpoint, with
every other parameter fixed and no further training. Balance edits rescale the two sides reciprocally.
Gain edits scale both sides together and then restore each matrix's Frobenius norm, so they hold the
global scale and the balanced core fixed and change only the scale fields (Appendix~\ref{app:gainedit}).

\begin{figure}[tbp]
\centering
\includegraphics[width=\textwidth]{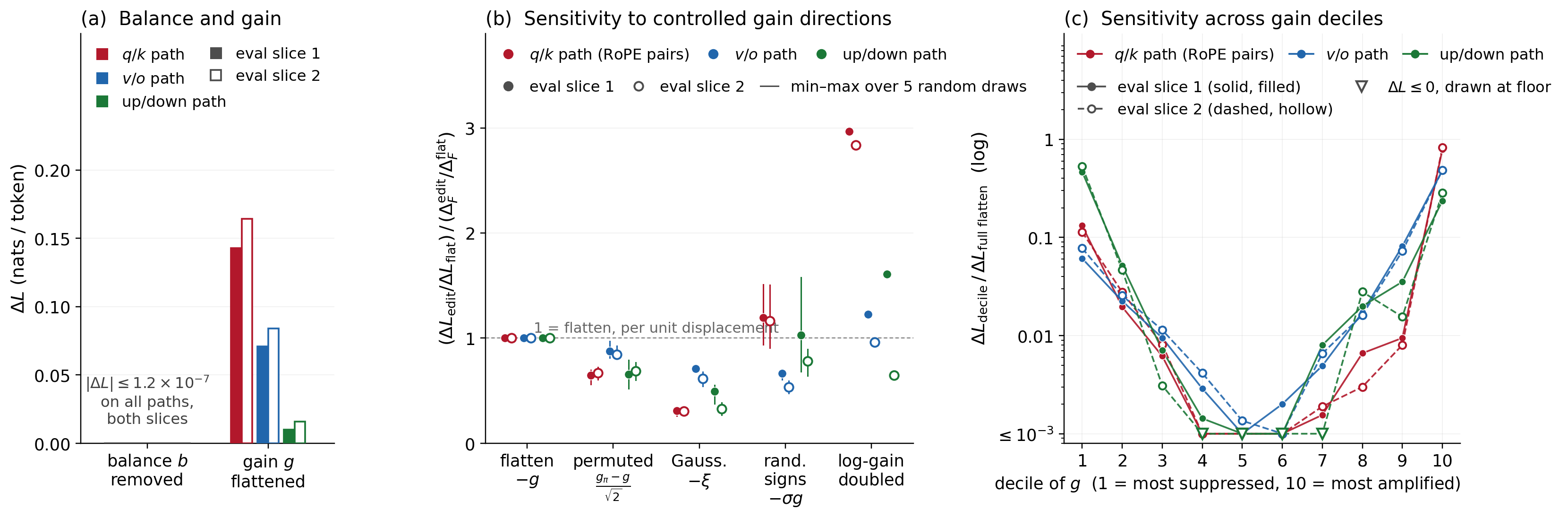}
\caption{\textbf{Frozen-checkpoint edits separate an invariant balance mode from a loss-sensitive gain
mode.} Edits act on one of the $q/k$, $v/o$ and up/down paths across all six layers of the
30{,}000-step base checkpoint. Filled and open markers show two disjoint evaluation slices from the training
token stream (slice 1 and slice 2; values in Table~\ref{tab:gainedit_matched}). \textbf{(a)} Balance removal
and gain flattening. \textbf{(b)} Loss changes relative to the flatten, divided by the corresponding ratios of
squared relative Frobenius displacement $\dF$. \textbf{(c)} Flattening one gain decile alone,
relative to the full flatten; responses below $10^{-3}$ of it are drawn at the floor, and the seven non-positive ones as open downward
triangles. Edit formulas and displacement ratios are in Appendix~\ref{app:gainedit}.}
\label{fig:gainedit}
\end{figure}

\paragraph{Balance preserves the computation; gain flattening changes the loss.}
Removing the balance on any path changes the loss only at the $10^{-7}$ level, the float32 resolution of
the mean loss, whereas flattening the gain raises it by 0.14, 0.07 and 0.01 nats per token (slice 1) on $q/k$, $v/o$
and up/down (Figure~\ref{fig:gainedit}a). Scale can move between the two matrices of a path without
changing the network; the relative channel gain is what the computation depends on, even at fixed matrix
norms and balanced core.

\paragraph{The trained gain direction costs more than random directions.}
Per unit squared displacement, flattening the trained gain costs 3.2--3.3, 1.4--1.6 and 2.0--3.1 times as much
as moving along a Gaussian direction in the same channel subspace on the two slices
(Figure~\ref{fig:gainedit}b). A displacement-matched permutation of the gain across channels costs roughly what its $1/\sqrt 2$ projection
onto the flatten direction plus a random remainder predicts (Appendix~\ref{app:gainedit}); this control
therefore gives no evidence of sensitivity to the channel assignment beyond that projection.

\paragraph{The response depends on path and direction.}
On $q/k$, doubling the log-gain profile ($\tau=-1$) costs 3.5--3.6 times as much as flattening it, at 1.2 times its
squared displacement; on $v/o$ and up/down doubling costs 0.6--1.6 times flattening. The response also concentrates
at the extremes of each field (Figure~\ref{fig:gainedit}c). Flattening only the most amplified decile
produces 0.8 of the full-field response on $q/k$ and 0.5 on $v/o$, whereas on up/down the most suppressed
decile produces 0.5 against 0.25 for the most amplified. After displacement normalization a concentration
remains at the amplified end on $q/k$ and at the suppressed end on up/down, but not on $v/o$. On up/down,
flattening the lowest-gain decile thus has the larger response under this intervention, which does not
define a general ranking of units.

The paired fields therefore supply functional edit coordinates. The ranges quoted above span both
evaluation slices. The evidence concerns finite edits of one checkpoint and seed,
scored on in-distribution loss; it does not show how the gain fields were generated or how such edits act
on training or generalization.

\section{Discussion}
\label{sec:discussion}

\paragraph{What the channel-scale description adds.}
The fields distinguish where magnitude is distributed across channels from the pooled magnitude
profile of a matrix. This distinction is useful even when pooled fits are similar: indexed fields
can differ across functional channels, and a narrow terminal field can follow substantial earlier
broadening. The complete representation $(s,h^r,h^c,Z)$ is invertible only when the full signed core
is retained (Appendix~\ref{app:conversions}). The two fields summarize channel RMS heterogeneity
with $d_{\mathrm{out}}+d_{\mathrm{in}}$ values; they do not replace the remaining matrix information.

\paragraph{From a common magnitude profile to different pooled shapes.}
The balanced cores have similar measured magnitude profiles across the tested settings, while
channel-scale heterogeneity accounts for much of the variation in pooled shape. The mixture bridge
makes this connection predictive on held-out runs and data arms within the controlled grid.
Its coefficients remain protocol- and setting-dependent, and equal field widths need not produce
equal pooled shapes: field distributions and scale--core pairing retain information beyond width
(Appendix~\ref{app:corebridge}). Likewise, the fitted scale tracks the global RMS through the
measured ratio $\lambda/s$, rather than being identical to $s$. These results explain why pooled
statistics can be informative while leaving channel organization unresolved.

\paragraph{Structural constraints and training-dependent outcomes.}
The computation graph identifies the channel spaces on which fields can be compared. The paired
RoPE reassignment further shows that the learned $q/k$ profiles follow assigned frequencies rather
than fixed row coordinates. Training determines the strength and trajectory of these profiles:
the sampled runs show continued $q/k$ broadening, partial $v/o$ recession and substantial FFN
narrowing. The extension to AdamW's second moment finds related functional organization using
log-space row and column main effects, a different read-out on the same parameter axes.
These observations constrain explanations of field formation but do not identify its generating
dynamics; neither the optimizer state nor the weight fields uniformly precede the other in the
available timing comparisons (Appendix~\ref{app:vfactors}).

\paragraph{Functional tests and prospective uses.}
The paired edits distinguish scale allocation between projections from their relative channel
gain. Reciprocal balance edits preserve the forward computation on the tested paths. Gain
flattening instead increases loss while preserving each matrix's global RMS and balanced core,
and exceeds the tested Gaussian controls after displacement normalization. The larger endpoint
response occurs at the high-gain end on $q/k$ and $v/o$, but at the low-gain end on up/down;
these are sensitivities to the specified finite edits, not a general ranking of unit importance.
The experiments demonstrate identity-aligned comparison, trajectory tracking and controlled
checkpoint edits. They motivate testing channel-aware rescaling or adaptation, but establish
neither gains in compression or generalization nor arrangement-specific sensitivity beyond the
reported controls.

\section{Limitations}
\label{sec:limitations}

\paragraph{Experimental coverage.}
The main controlled trajectories and interventions use a LLaMA-style 70M model and one corpus with
transformed data conditions. The separately trained Pythia size series provides descriptive single-seed
comparisons, while public Pythia checkpoints extend static comparisons to four sizes without optimizer or
intervention records. Optimizer-state records, the no-position control and the checkpoint edits each use one
seed; the RoPE reassignment uses three paired seeds. Three initialization families support the qualitative
trajectory comparison, with the detailed field-width analysis centred on Gaussian runs. Matrices and layers
within a run are not independent replications, and aggregate profiles can conceal depth-dependent
differences, as observed for learned channel multipliers \citep{velikanov2026multipliers}. Transport to substantially larger models, other corpora and
other optimizers is not established.

\paragraph{Measurement and statistical summaries.}
The core is defined by two-sided RMS balancing; its fitted shape is measured with a middle-80\% magnitude
protocol, supplemented by the displayed quantile comparisons. Similar measured profiles do not imply
identical cores, independent entries, Gaussianity or universal tail behaviour. The bridge is a calibrated
empirical relation with setting-dependent coefficients and a residual scale--shape pairing term; field widths
are not sufficient statistics for the pooled shape. The tracking of $s$ by $\lambda$ rests on the measured
stability of the ratio $\lambda/s$, not on their equality. The $q/k$ pair profiles aggregate weight-field
read-outs, robust to the choice between the mean of the two rows' log RMS and their pooled RMS, whereas the optimizer factors
are row and column main effects of $\log \hat V$.

\paragraph{Generating dynamics.}
The RoPE intervention changes the assignment of a fixed frequency set. It identifies frequency-based
organization, not the effect of changing the spectrum or the origin of the low-frequency preference; rotary
symmetry makes the pair the natural unit of comparison but does not determine the learned coupling strength.
Optimizer-state correspondence and timing remain observations inside a coupled training loop and do not
establish how the different field trajectories are generated, including why the attention-side fields are
retained while the FFN fields recede.

\paragraph{Functional and practical scope.}
The edits are finite perturbations of one checkpoint, evaluated on two disjoint slices of the training
token stream. They test forward-pass invariance and in-distribution loss sensitivity, not subsequent training
behaviour, held-out task performance or generalization. Control edits are compared after displacement
normalization rather than at exactly equal displacements, and the permutation comparison uses a
quadratic-response approximation. The decile responses do not define a pruning criterion, and benefits for
quantization, channel selection or scale-only adaptation remain untested. Weight-side read-outs also do not
measure learning quality on their own: the companion study finds data conditions with similar weight-side
read-outs that differ widely in generalization.

\section{Conclusion}

Row and column scale fields provide a mesoscopic, channel-indexed description between individual
Transformer weights and pooled matrix statistics. Together with the global RMS and the full balanced core
they retain the matrix information, while their widths and profiles summarize channel heterogeneity and
organization. In the tested settings, similar balanced-core magnitude profiles coexist with different pooled
shapes, much of whose departure is predicted by field width.

The fields align with functional channel spaces and evolve differently across components during training.
The same channel-based analysis extends to AdamW's second-moment factors, and frozen-checkpoint edits
distinguish an invariant reciprocal balance from a loss-sensitive gain. These results establish a framework
for comparing, tracking and experimentally probing channel-scale structure; its generating dynamics, its
transport to other settings and its benefits for model adaptation or generalization are questions for
further study.

\bibliography{references}

\clearpage
\appendix
\section{Methods and reproducibility}
\label{app:protocol}

This appendix describes the decomposition, the read-outs and the training runs.
Table~\ref{tab:configs} states the protocol and the role of each evidence source.

\subsection{Notation}
\label{app:notation}

Table~\ref{tab:notation} lists the symbols used across sections, with the equation or section that defines
them; symbols local to one appendix are defined where they appear.
Throughout, $\log$ denotes the natural logarithm; $\log_{10}$ is used only for the histogram grid and for
$H^W_{\mathrm{IQR}}$. Profile correlations $r$ are Pearson correlations unless stated otherwise.

\begin{table}[p]
\centering\scriptsize
\setlength{\tabcolsep}{4pt}
\begin{tabularx}{\textwidth}{@{}l X l@{}}
\toprule
symbol & meaning & defined in \\
\midrule
\multicolumn{3}{@{}l}{\textbf{Matrix and decomposition}} \\
$W \in \mathbb{R}^{d_{\mathrm{out}} \times d_{\mathrm{in}}}$ & one projection matrix; rows index output channels, columns input channels & Section~\ref{sec:decomp} \\
$W = s\,D_r\,Z\,D_c$ & exact re-parameterization: global scale, row and column balancing factors, core & eq.~\eqref{eq:decomp} \\
$s = \mathrm{RMS}(W)$ & global scale, fitting-free & eq.~\eqref{eq:decomp} \\
$D_r,\,D_c$; $(D_r)_{ii},\,(D_c)_{jj}$ & positive diagonal balancing factors; convertible to and from the scale fields given $s$ and $Z$ & eqs.~\eqref{eq:decomp}, \eqref{eq:recover} \\
$Z$ & two-sided RMS-balanced core, unit RMS in every row and column & eq.~\eqref{eq:balanced_core} \\
$Z^r$ & one-sided (row-normalized) core, $W = \mathrm{diag}(R)\,Z^r$; its shape is $\krow$ & eq.~\eqref{eq:onesided} \\
$\chi^r_i,\,\chi^c_j$ & log $Z^2$-weighted RMS of the opposite-side factors; separates $h^r$ from $\log D_r$ (and $h^c$ from $\log D_c$) & Appendix~\ref{app:conversions} \\
$\mathcal C(\cdot)$ & median-centring over channels, $\mathcal C(x) = x - \operatorname{median}(x)$ & Appendix~\ref{app:conversions} \\
$R_i = \mathrm{RMS}_j(W_{ij})$, $C_j = \mathrm{RMS}_i(W_{ij})$ & row and column (channel) RMS of $W$ & eq.~\eqref{eq:fields} \\
\midrule
\multicolumn{3}{@{}l}{\textbf{Scale fields}} \\
$h^r_i,\,h^c_j$ & row and column scale fields: median-centred log row and column RMS & eq.~\eqref{eq:fields} \\
$h$ & either field when the axis is clear; $h_1, h_2$ the identity-axis fields of two paired projections & Section~\ref{sec:gainedit} \\
$\Hrow,\,\Hcol$ & field widths $\mathrm{sd}_i(h^r_i)$, $\mathrm{sd}_j(h^c_j)$ & Appendix~\ref{app:fitting} \\
$\HW$ & field width (standard deviation) on a kind's identity side (Table~\ref{tab:paths}) & Section~\ref{sec:decomp} \\
$H^W_{\mathrm{IQR}}$ & interquartile row width used for the RoPE intervention & Appendix~\ref{app:fitting} \\
\midrule
\multicolumn{3}{@{}l}{\textbf{Weibull read-outs}} \\
shape $k$ & Weibull shape from the middle-80\% probability-plot fit & Appendix~\ref{app:fitting} \\
$\kraw$ & shape of the original matrix (pooled shape) & Appendix~\ref{app:fitting} \\
$\krow,\,\kcol$ & shape after row or column normalization & Appendix~\ref{app:fitting} \\
$\kident$ & $\krow$ or $\kcol$ on the identity side & eq.~\eqref{eq:bridge} \\
$\kbi$ & shape of the two-sided core $Z$ & Appendix~\ref{app:fitting} \\
$\lambda$, $\lambda_{\mathcal P}$ & fitted Weibull scale; the scale returned by protocol $\mathcal P$ & eq.~\eqref{eq:lambdaequiv} \\
$\dkmix$ & one-axis mixture term $\kraw^{-2} - \kident^{-2}$ & eq.~\eqref{eq:bridge} \\
$\alpha_P$; $\alpha_r,\,\alpha_c$ & protocol coefficient of the one-axis bridge; row and column coefficients of the two-axis bridge & eqs.~\eqref{eq:bridge}, \eqref{eq:bridge2} \\
$6/\pi^2$ & reference coefficient for an exact Weibull under independence & Appendix~\ref{app:weibullref} \\
$R_0^2$, $R^2$ & through-origin and centred coefficients of determination & Appendix~\ref{app:fitting} \\
$r$; $\rho$ & profile correlation; Spearman rank correlation & Appendix~\ref{app:fitting}, \ref{app:vfactors} \\
$r_{\mathrm{freq}},\,r_{\mathrm{coord}},\,\Delta r_{\mathrm{id}}$ & paired-profile correlation by assigned frequency, by matrix coordinate, and their difference & Section~\ref{sec:identity_rope} \\
\midrule
\multicolumn{3}{@{}l}{\textbf{Optimizer state}} \\
$G_t$ & mini-batch gradient of a projection, a sum of per-token outer products $\delta x^{\top}$ & eq.~\eqref{eq:adamw_state} \\
$M_t,\,V_t$; $\hat M_t,\,\hat V_t$ & AdamW first and second moments; bias-corrected & eqs.~\eqref{eq:adamw_state}, \eqref{eq:adamw_update} \\
$U_t$, $\epsilon$, $\eta_t$, $\gamma_{\mathrm{wd}}$, $\beta_1,\,\beta_2$ & update direction, stabilizer, learning rate, decoupled weight decay, EMA coefficients & eq.~\eqref{eq:adamw_update} \\
$\log \hat V_{ij} = \mu + a_i + b_j + \varepsilon_{ij}$ & additive row--column model of the second moment; $\varepsilon_{ij}$ the coordinate remainder & eq.~\eqref{eq:vfactors} \\
\midrule
\multicolumn{3}{@{}l}{\textbf{Data arms and checkpoint edits}} \\
$g = h_1 + h_2$, $b = h_1 - h_2$ & path gain and balance modes of a paired identity axis & Section~\ref{sec:gainedit} \\
$\tau$ & strength of a gain edit ($\tau = 1$ flattens the gain, $\tau = -1$ doubles it) & Appendix~\ref{app:gainedit} \\
$\Delta L$ & change in mean next-token loss (nats per token) after an edit & Appendix~\ref{app:gainedit} \\
$\dF$, $\dF^{\mathrm{flat}}$ & relative squared Frobenius displacement of an edit; that of the flatten at $\tau = 1$ & Appendix~\ref{app:gainedit} \\
\bottomrule
\end{tabularx}
\caption{Notation. Component kinds are written $q$, $k$, $v$, $o$, gate, up and down; the key projection $k$ always appears alongside $q$, and the Weibull shape is always qualified as ``shape $k$''.}
\label{tab:notation}
\end{table}

\subsection{Conversions between representations}
\label{app:conversions}

Equation~\eqref{eq:chain} consists of two links, the second at fixed $s$ and $Z$:
\begin{equation*}
W \;\overset{\text{(i)}}{\longleftrightarrow}\; (s,\,D_r,\,Z,\,D_c) \;\overset{\text{(ii), (iii)}}{\longleftrightarrow}\; (s,\,h^r,\,h^c,\,Z).
\end{equation*}
Link (i) is the balancing decomposition; link (ii) converts the balancing factors into the fields, (iii)
inverts it, and (iv) shows that the inversion recovers $W$ uniquely. Throughout, the matrix is entrywise
nonzero, as every self-trained matrix is (the public Pythia checkpoints contain exact zeros in about
$10^{-6}$ of their entries, for which the balancing also converges), the core $Z$ is kept as the full signed matrix, $\log D_r$ and $\log D_c$ denote the vectors of log
diagonal entries, and $\mathcal C(x) = x - \operatorname{median}(x)$ denotes median-centring.

\paragraph{(i) Matrix and balancing factors, $W \leftrightarrow (s, D_r, Z, D_c)$.} Balancing $W$ gives the
factors of eq.~\eqref{eq:decomp}, and their product returns $W$. Condition~\eqref{eq:balanced_core} is a
matrix-scaling problem on the entrywise squares $W \circ W$, solved by alternating row and column RMS
normalization; its solution is unique up to $(D_r, D_c) \to (cD_r, D_c/c)$ (see (iv)). The constant $c$ is fixed by the
iteration itself, in which both factors start at one and rows are normalized before columns in each sweep;
the fields, their widths and $Z$ do not depend on it. The iteration stops at a relative row- and column-RMS
spread of $10^{-6}$ (at most 200 sweeps), and every reported self-trained matrix passes the checks spread
$<10^{-3}$ and reconstruction $W = s\,D_r Z D_c$ within $10^{-9}$ of the global RMS. The public Pythia
matrices and two early extraction scripts used a looser routine (60 sweeps, tolerance $10^{-5}$);
re-balancing the 348 public Pythia matrices with the routine above changes $\kbi$ by at most $10^{-6}$.

\paragraph{(ii) Factors to fields, $(D_r, D_c) \rightarrow (h^r, h^c)$.} Substituting eq.~\eqref{eq:decomp}
into the row RMS of eq.~\eqref{eq:fields} gives $\mathrm{RMS}_j(W_{ij}) = s\,(D_r)_{ii}\,e^{\chi^r_i}$, so
\begin{equation}
\begin{aligned}
h^r &= \mathcal C(\log D_r + \chi^r), &\qquad
\chi^r_i &= \tfrac12 \log\Big(\tfrac{1}{d_{\mathrm{in}}} \textstyle\sum_j Z_{ij}^2\,(D_c)_{jj}^2\Big),\\
h^c &= \mathcal C(\log D_c + \chi^c), &\qquad
\chi^c_j &= \tfrac12 \log\Big(\tfrac{1}{d_{\mathrm{out}}} \textstyle\sum_i Z_{ij}^2\,(D_r)_{ii}^2\Big).
\end{aligned}
\label{eq:factorsfields}
\end{equation}
Each field thus carries its own factor and a term $\chi$ from the opposite factor. Since the rows of $Z$ have
unit RMS (eq.~\eqref{eq:balanced_core}), $e^{2\chi^r_i}$ is a $Z^2$-weighted mean of $(D_c)_{jj}^2$ along row $i$,
and $h^r = \mathcal C(\log D_r)$ holds exactly when $\chi^r$ is constant across rows, for instance when $D_c$ is
a multiple of the identity. This is not the case in general, so the fields and the centred log factors are
related exactly but need not coincide entrywise. Median centring is not additive: $\mathcal C(\log D_r + \chi^r)$
and $\mathcal C(\log D_r) + \mathcal C(\chi^r)$ differ by a constant, which leaves widths and correlations unchanged.

\paragraph{(iii) Fields to factors, $(h^r, h^c) \rightarrow (D_r, D_c)$.} Solving eq.~\eqref{eq:factorsfields}
for the factors, with the unknown centring constants fixed by $\mathrm{RMS}(D_r Z D_c) = 1$, gives
\begin{equation}
\log D_r = h^r - \chi^r - \tfrac12 \log\Big(\tfrac{1}{d_{\mathrm{out}}} \textstyle\sum_{i} e^{2h^r_{i}}\Big), \qquad
\log D_c = h^c - \chi^c - \tfrac12 \log\Big(\tfrac{1}{d_{\mathrm{in}}} \textstyle\sum_{j} e^{2h^c_{j}}\Big).
\label{eq:recover}
\end{equation}
Because $\chi^r$ depends on $D_c$ and $\chi^c$ on $D_r$, the two equations are solved by updating them in turn,
which is the balancing scheme of (i) with row and column targets set by the fields.
The global scale does not enter; eq.~\eqref{eq:decomp} then returns $W = s\,D_r Z D_c$.

\paragraph{(iv) Uniqueness.} At a fixed point of eq.~\eqref{eq:recover} the matrix
$\Omega = D_r^2\,(Z \circ Z)\,D_c^2$, with $Z \circ Z$ the entrywise square, has row sums
$d_{\mathrm{in}} e^{2h^r_i}$ and column sums $d_{\mathrm{out}} e^{2h^c_j}$, each divided by the channel mean of
$e^{2h}$ on its side as in eq.~\eqref{eq:recover}, so that both totals equal $d_{\mathrm{out}} d_{\mathrm{in}}$. For a positive matrix the diagonally scaled
matrix with prescribed row and column sums is unique \citep{sinkhorn1967prescribed}, which covers the
self-trained matrices since they have no exact zeros. By (ii), $(W \circ W)/s^2$
has this form and these sums, so $\Omega = (W \circ W)/s^2$ at any fixed point: the factors are determined up to
$(D_r, D_c) \to (cD_r, D_c/c)$, and $W_{ij} = s\,\operatorname{sgn}(Z_{ij})\,\Omega_{ij}^{1/2}$, the positive factors
preserving signs.

The conversions need the full signed core with its channel indices. Field widths and pooled magnitude
profiles are summaries, and the relation between them is the empirical mixture bridge of
Section~\ref{sec:bridgedef}.

\subsection{Training configurations}
\label{app:configs}

All LLaMA-style \citep{touvron2023llama} runs train the same 70M-parameter decoder (6 layers, width 512, FFN
width 1{,}376, 8 heads of dimension 64, SwiGLU \citep{shazeer2020glu}, RoPE \citep{su2024roformer}, RMSNorm
\citep{zhang2019rmsnorm} with $\epsilon = 10^{-5}$, untied output layer, initialization
scale 0.02) in fp32 with AdamW \citep{kingma2015adam, loshchilov2019decoupled} ($\beta_1 = 0.9$, $\beta_2 = 0.999$, $\epsilon = 10^{-8}$, decoupled weight decay 0.1
applied to all parameters, including embeddings, the output layer and the RMSNorm gains) and no gradient
clipping, at peak learning rate $10^{-3}$ with 200 warm-up steps and a cosine schedule of a 30{,}000-step
horizon that decays to 10\% of peak, batch size 24 and sequence length 512 unless Table~\ref{tab:configs}
states otherwise. The corpus is the first $10^8$ tokens of the WikiText-103 training split with empty lines
removed \citep{merity2017pointer}, tokenized with the Pythia (GPT-NeoX) tokenizer (vocabulary padded to
50{,}304). Two sampling schemes are used. The controlled grid reads a transformed stream sequentially in a
single pass: for a token budget $T$, the first $T/n_{\mathrm{rep}}$ tokens form its unique block, a fraction $f$ of
the positions in that block have their tokens permuted among themselves, and the block is tiled
$n_{\mathrm{rep}}$ times. All 30{,}000-step runs instead sample 512-token windows uniformly with replacement from
the $10^8$-token stream, about 3.7 expected passes. Weight read-outs come from stored checkpoints;
optimizer-state read-outs and field-edit evaluations use their own stored states and evaluation streams.
Matrix fits are nested within runs.

\paragraph{Protocols that differ.}
In the paired RoPE reassignment, the base of pair 1 repeats the seed-1 Gaussian initialization run (same
initialization and batch sequence), and pairs 2--3 use the seed-2 and seed-3 Gaussian runs as base arms and
are compared at the three checkpoints 3{,}200, 10{,}000 and 30{,}000 that both arms store. The controlled grid
uses a token budget of $T = 61.44$M, is checkpointed at steps 800, 1{,}600, 3{,}200 and
5{,}000, and stops at 5{,}000 steps of the 30{,}000-step schedule, at about 94\% of the peak learning rate. The
optimizer-state records come from two runs. A dump run in the seed-1 Gaussian configuration (batch 24, same
batch sequence as the base run) stops at 10{,}000 steps of the 30{,}000-step schedule and stores the full $W$,
$\hat M$ and $\hat V$ of every projection at steps 200, 400, 800, 3{,}200 and 10{,}000 (Figure~\ref{fig:topology}
and the residual comparison of Appendix~\ref{app:vfactors}). A separate full-length Gaussian run with batch
size 16 and its own window sequence logs axis-wise $W$, $\hat M$ and $\hat V$ statistics over a 20-step window
every 200 steps (lock-in timing of Appendix~\ref{app:vfactors}). The Pythia size series uses batch
$24 \times 512$, 200 warm-up steps and cosine decay to 10\% of peak over its 8{,}000-step budget, with
checkpoints at 512, 2{,}000, 5{,}000 and 8{,}000; its windows are sampled with replacement from a 98.3M-token
arm stream built from the full 118.7M-token WikiText-103 training split, pre-tokenized separately.

\begin{table}[tbp]
\centering\footnotesize
\setlength{\tabcolsep}{4pt}
\renewcommand{\arraystretch}{1.15}
\begin{tabularx}{\textwidth}{@{}>{\raggedright\arraybackslash}p{2.4cm} >{\raggedright\arraybackslash}p{2.6cm} >{\raggedright\arraybackslash}p{1.5cm} >{\raggedright\arraybackslash}p{1.8cm} >{\raggedright\arraybackslash}p{1.6cm} >{\raggedright\arraybackslash}X@{}}
\toprule
source & role & model & runs & steps / ckpts & protocol-specific details \\
\midrule
Controlled data grid & one-axis bridge; short trajectories & LLaMA 70M & 4 arms $\times$ 3 seeds & 5{,}000 / 4 & shuffling $f\in\{0,1\}$ $\times$ repetition $n_{\mathrm{rep}}\in\{1,64\}$: D1 $(f{=}0,n_{\mathrm{rep}}{=}1)$, D2 $(1,1)$, D3 $(1,64)$, D4 $(0,64)$; sequential single pass; stopped at 5{,}000 steps of the 30{,}000-step schedule. \\
Paired RoPE reassignment & frequency-to-row identity (paired intervention) & LLaMA 70M & 3 seed pairs & 30{,}000 / 9; 3 shared (pairs 2--3) & base run against a run whose RoPE inverse-frequency table is permuted by a fixed $\pi$ at initialization; frequency set unchanged; base arms in the text. \\
Initialization families & core convergence; long-term field evolution & LLaMA 70M & 3 families $\times$ 3 seeds & 30{,}000 / 7 (0, 800, 3{,}200, 10k, 20k, 25k, 30k) & Gaussian, Laplace, uniform at equal initial RMS; windows sampled with replacement from the untransformed $10^8$-token stream. \\
Optimizer-state records & row--column structure of $\hat V$ (descriptive) & LLaMA 70M & 2 & 10{,}000 / 5 dumps; 30{,}000 / 150 logged windows & a dump run (batch 24) and a separate logger run (batch 16); see text. \\
No-position control & RoPE-versus-head organization (boundary) & LLaMA 70M & 1 & 30{,}000 / 9 (6 shared with the base run) & base run with the rotary rotation removed ($\cos=1$, $\sin=0$); single-seed boundary. \\
Checkpoint field edits & function carried by the paired gain (screen) & LLaMA 70M & base run at 30k & no training & edits of the stored weights scored by mean loss on two disjoint slices of 48 sequences $\times$ 512 tokens from the training stream; 63 edits and 54 displacement-controlled edits (Appendix~\ref{app:edits}). \\
Pythia size series & transport across model size (descriptive) & Pythia 70M / 160M / 410M & 4 arms $\times$ 3 sizes, 1 seed & 8{,}000 / 4 & the four grid arms at three sizes under the companion study's Pythia protocol: GPT-NeoX architecture \citep{black2022gptneox} (FFN 2{,}048 at 70M, rotary on 25\% of head dimensions, fused QKV), its default initialization, weight decay 0.01, peak learning rate $3\times10^{-4}$; schedule and corpus in the text. \\
Public Pythia endpoints & static component and size comparison (descriptive) & Pythia 70M--1B & 4 sizes, 1 model each & final checkpoint of each & released final checkpoints \citep{biderman2023pythia}, refitted raw and after identity-axis normalization; no gate projection, so Figure~\ref{fig:core}(a) shows $q$, $k$, $v$ (split from the fused QKV), $o$ and the two FFN projections. \\
\bottomrule
\end{tabularx}
\caption{Evidence sources, their roles and protocol-specific details; all LLaMA-style runs share the architecture, optimizer and tokenizer stated above, while sampling, schedule length and batch size differ by row. Rows denote evidence streams rather than independent sample counts: the Gaussian initialization runs are the base arms for RoPE seed pairs 2 and 3. Fit counts and the mapping to figures are in the data documentation.}
\label{tab:configs}
\end{table}

\subsection{Weibull fitting protocol and the spread statistic}
\label{app:fitting}

We use the middle-80\% probability-plot fit of \citet{ding2026a_weibull}, implemented on a histogram: $|W|$
is binned into a 1{,}024-bin $\log_{10}$ histogram on $[-12, 2]$ and fitted by a probability-plot regression of
$\log(-\log(1-F))$ on $\log x$ at the bin centres $x$, where $F$ is the empirical cumulative distribution of $|W|$
evaluated at the mid-point of each bin (for a Weibull distribution this plot is linear with slope $k$);
the regression is weighted by bin count and restricted to the bins with $F$ in the middle 80\%; the slope is the shape $k$ and the intercept gives
$\lambda = \exp(-c_0/k)$ with $c_0$ the intercept. The same protocol is applied to the original, row-normalized, column-normalized
and two-sided balanced matrices, giving $\kraw$, $\krow$, $\kcol$ and $\kbi$; unless stated otherwise
$\lambda$ is the fitted scale of the original matrix. The $R^2$ of the regression is recorded for every
fit with a gate of $0.99$; matrices with a fit below the gate are retained and flagged, and figures that
apply the gate say so. They occur only in the uniform initialization family at steps 0 and 800 (133 of 2{,}646
matrix records) and in two raw $q$ fits of the public Pythia endpoints, so the gate removes no Gaussian-family
matrix. No entry of the audited self-trained checkpoints lies below $10^{-12}$ and none is exactly zero; the
public Pythia checkpoints contain 1{,}326 exact zeros (about $10^{-6}$ of their entries), which fall outside the
grid and are excluded. Every coefficient relating $\HW$ to the shape $k$ is a coefficient of this window and
estimator.

The field widths are the population standard deviations (normalized by the channel count) of the
natural-log fields, $\Hrow = \mathrm{sd}_i(h^r_i)$ and $\Hcol = \mathrm{sd}_j(h^c_j)$, and $\HW$
without a subscript denotes the width on a kind's identity axis (Table~\ref{tab:paths}). The paired RoPE comparison is read on
the row side for every kind with a second, interquartile statistic fixed before the intervention,
$H^W_{\mathrm{IQR}} = \mathrm{IQR}_i\big(\log_{10} \mathrm{RMS}_j(W_{ij})^2\big)$, which weights the central rows where
$\mathrm{sd}$ weights the tails; the two can move differently under the permutation (Appendix~\ref{app:axes}).

\paragraph{Bridge statistics.}
All bridge coefficients are fitted through the origin and scored by the through-origin
$R_0^2 = 1 - \sum_n (y_n - \widehat y_n)^2 / \sum_n y_n^2$ over the fitted matrices $n$, with $\widehat y_n = \alpha_P x_{r,n}$ for the
one-axis model and $\widehat y_n = \alpha_r x_{r,n} + \alpha_c x_{c,n}$ for the two-axis model; this is not
the centred $R^2$. The statistical unit is the training run (one seed at one data condition, an arm).
Leave-one-run-out refits the coefficient on the other runs and predicts the held-out run;
leave-one-data-arm-out refits on three arms and predicts the fourth.

\paragraph{Global scale and fitted scale.}
\label{app:lambdaequiv}
The global scale $s = \mathrm{RMS}(W)$ is computed without fitting and scales in proportion to $W$. For an
ideal probability-plot fit the fitted scale $\lambda$ does the same, because a common factor shifts the plot
sideways without changing its slope. With the fixed log grid, a shift that is not a whole number of bins
(0.0137 decades) changes the binning slightly: refitting $W/s$ for the 378 matrices of the base run of the
paired RoPE reassignment reproduces $\lambda/s$ within 0.18\% and $k$ within 0.0040 (medians 0.06\% and 0.001),
which is the approximate form of eq.~\eqref{eq:lambdaequiv}. On that run (42 matrices, nine stored steps),
$\log s$ rises over training by 1.26 on $q$ and $k$, 1.04 on gate, 0.94 on up and down, and 0.74--0.77 on $v$
and $o$, while the layer medians of $\lambda/s$ stay within 0.845--0.887 on every kind (individual matrices
0.786--0.890), so $\lambda$ follows the global scale. The upper end matches the value $\lambda/s \approx 0.8875$
derived for a Gaussian initialization under the middle-80\% protocol \citep{ding2026a_weibull}. The measured ratio lies 3--9\% above the Weibull value
0.817 of eq.~\eqref{eq:weibullratio} at $k \approx 1.20$, comparable to the $\approx$4.6\% bridge residual between the
RMS-derived and the fitted $\lambda$ that \citet{ding2026b_adamw} attribute to the nonlinearity of the Weibull fit.

\subsection{The full-Weibull reference coefficient}
\label{app:weibullref}

For $X \sim \mathrm{Weibull}(k, \lambda)$, $E = (X/\lambda)^k$ is standard exponential, so
$\log X = \log\lambda + k^{-1}\log E$ and $\mathrm{Var}(\log X) = k^{-2}\,\mathrm{Var}(\log E)$. The
moments $\mathbb{E}[E^{m}] = \Gamma(m+1)$ give $\mathrm{Var}(\log E) = \psi'(1) = \sum_{n \ge 1} n^{-2} =
\pi^2/6$, with $\psi$ the digamma function (equivalently, $-\log E$ is standard Gumbel). Hence
$k^{-2} = (6/\pi^2)\,\mathrm{Var}(\log X)$. For the one-sided decomposition $W_{ij} = \mathrm{RMS}_j(W_{ij})\, Z^r_{ij}$,
pooling all entries uniformly gives exactly $\mathrm{Var}(\log|W|) = (\Hrow)^2 +
\mathrm{Var}(\log|Z^r|) + 2\,\mathrm{Cov}_i\big(h^r_i, \mathrm{mean}_j \log|Z^r_{ij}|\big)$. Because every row of $Z^r$ has unit
RMS, the row mean of $\log|Z^r_{ij}|$ is a statistic of the within-row shape, so the last term pairs each row's scale
with its shape; it is the log-variance counterpart of the pairing term of Appendix~\ref{app:corebridge}.
Neglecting it (approximate independence of the row scale and $Z^r$) and applying the log-variance identity to the raw and the row-normalized
magnitudes gives
\begin{equation}
\kraw^{-2} - \krow^{-2} \approx \frac{6}{\pi^2}\,(\Hrow)^2 ,
\end{equation}
with equality when both the raw and the row-normalized magnitudes are exactly Weibull and the pairing term
vanishes; the column case follows by transposition. $6/\pi^2$ is
therefore the reference coefficient of the mixture bridge under those two conditions, not a value that
the middle-80\% protocol coefficient $\alpha_P$ must attain.

\subsection{Code and data}
\label{app:codedata}

The release is the directory \texttt{scale\_field} at tag \texttt{p5-scale-field-v1} of the repository
\begin{center}\url{https://github.com/tiexinding/NPM-Weibull-public}\end{center}
It contains the training code and configurations of the self-trained runs, the script that builds the
pre-tokenized training stream from the public corpus, the analysis code that computes the data tables from
the checkpoints and optimizer states and the reported statistics from the tables, the data tables, and
documentation of their contents. Model checkpoints and optimizer dumps are not released because of their
size.

\clearpage
\section{Supporting results}
\label{app:supporting}

The supporting results follow the order of Section~\ref{sec:results}: the common core, the mixture
bridge, the functional axes and RoPE, the weight-field trajectories, the AdamW second-moment factors, and
the paired gain modes and frozen-checkpoint edits.

\subsection{Common core}
\label{app:commoncore}

This appendix bounds the common-core statement of Section~\ref{sec:core}. Figure~\ref{fig:core} reports component-level medians, not an identical core in every matrix. Across
the 2{,}016 two-sided fits of the controlled grid the individual $\kbi$ values span 1.18--1.27, with the
upper end in the two shuffled arms. The $v$ projection of the shuffled high-repetition arm shows the
largest one-sided departure: two-sided balancing moves its lowest block from $\krow = 1.09$ to
$\kbi = 1.19$. Among the normalized public-Pythia blocks, a few $q$ blocks overshoot the reference (up to 1.28 at 410M)
and a few $k$ blocks fall below it (down to 1.125 at 70M), although the block medians of both kinds lie in
1.193--1.205. Per-kind and per-checkpoint values are in the data
package.

\subsection{Mixture bridge}
\label{app:corebridge}

This appendix supports the bridge of Section~\ref{sec:bridge} and the axis choice of
Section~\ref{sec:identity}: hold-out prediction, axis-specific normalization, the two-axis form, and what
the width-only bridge leaves out.
Fits are nested in their training runs and are never treated as independent samples.

To check that the bridge of Figure~\ref{fig:bridge} is not carried by many matrices of the same
training trajectory, its coefficient is refitted with a whole run or a whole data arm withheld
(definitions in Appendix~\ref{app:protocol}). The held-out errors stay small on average, and the
shuffled high-repetition arm D3 is the worst fold for every kind under both schemes
(Figure~\ref{fig:bridge}d; absolute leave-one-run-out errors in Figure~\ref{fig:twoaxis}c). The
coefficient is positive in all 12 runs for the five row-side kinds (Figure~\ref{fig:twoaxis}a), while
its value moves from 0.791 on the grid to 0.654 on pooled $q/k$ of the paired RoPE-reassignment runs
and drifts in time by about a factor 1.3 on the FFN side. The relation is therefore predictive within
the grid; its numerical coefficient is a property of the protocol and the setting.

\subsubsection{\texorpdfstring{Axis-specific normalization and the two-sided $v$ exception}{Axis-specific normalization and the two-sided v exception}}

Table~\ref{tab:sides} gives, per kind, the raw fit, the fits after row, column and two-sided
normalization, and the through-origin $R_0^2$ on each axis. Read along a row, the side whose
normalization returns the fit to the reference band and whose width explains the departure is the
identity side of Table~\ref{tab:paths}. Which
side carries the identity is fixed by the computation graph and tested by the cross-projection pairing of
Section~\ref{sec:identity}; which side's measured field is the wider one usually follows, but that is
empirical: it holds throughout training for $q$, $k$ and $o$, the two sides stay comparable for $v$, and
in the FFN kinds, whose fields are the narrowest, the wider side is not stable over training. $v$ is the
exception: removing only one side leaves the other field mixed into the pooled distribution, which is why
$v$ alone needs two-sided normalization to recover the common core. An independent check on the paired
RoPE-reassignment runs gives the same direction: the pooled shape $k$ of $q/k$ is returned to the initialization
value by row and not by column normalization, while that of $W_o$ is returned by column and not by row
normalization.

\begin{table}[tbp]
\centering\small
\begin{tabular}{@{}l cccc c c l@{}}
\toprule
kind & $\kraw$ & $\krow$ & $\kcol$ & $\kbi$ & $\Hrow$/$\Hcol$ & $R_0^2$ row/col/both & identity side (Table~\ref{tab:paths}) \\
\midrule
$q$ & 1.192 & 1.208 & 1.195 & 1.211 & 0.16 / 0.06 & 0.978 / 0.623 / 0.978 & rows \\
$k$ & 1.188 & 1.207 & 1.192 & 1.210 & 0.17 / 0.08 & 0.977 / 0.657 / 0.977 & rows \\
$v$ & 1.194 & 1.200 & 1.197 & 1.205 & 0.13 / 0.11 & 0.932 / 0.942 / 0.964 & both \\
$o$ & 1.191 & 1.195 & 1.199 & 1.206 & 0.12 / 0.16 & 0.799 / 0.925 / 0.927 & columns \\
gate & 1.187 & 1.203 & 1.188 & 1.205 & 0.19 / 0.07 & 0.906 / 0.627 / 0.915 & rows \\
up & 1.191 & 1.203 & 1.192 & 1.206 & 0.18 / 0.08 & 0.890 / 0.726 / 0.893 & rows \\
down & 1.186 & 1.186 & 1.201 & 1.204 & 0.09 / 0.20 & 0.431 / 0.878 / 0.926 & columns \\
\bottomrule
\end{tabular}

\caption{Medians over the 288 fits per kind on the controlled grid (12 runs $\times$ 4 checkpoints
$\times$ 6 layers): the raw fit, the fit after row, column and two-sided normalization, the two
heterogeneities, and $R_0^2$ of the through-origin bridge $\kraw^{-2} - \kbi^{-2}$ against
$(\Hrow)^2$, $(\Hcol)^2$, and both.}
\label{tab:sides}
\end{table}

\subsubsection{The two-axis form of the bridge and its identifiability}

\begin{figure}[tbp]
\centering
\includegraphics[width=\textwidth]{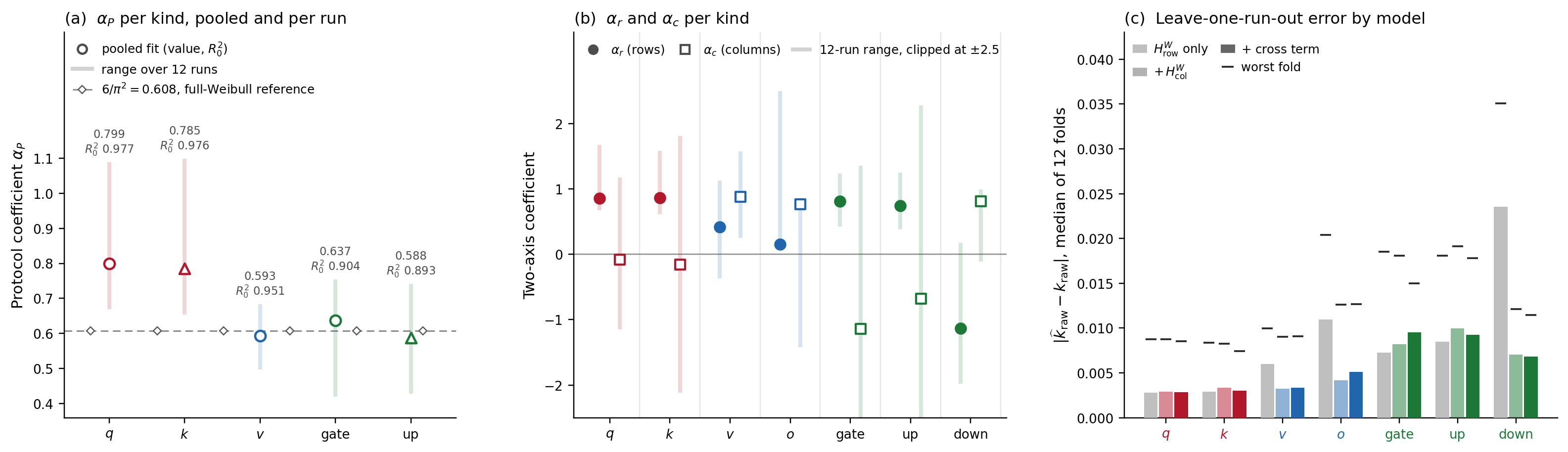}
\caption{\textbf{Bridge coefficients per kind, on one axis and on two.}
\textbf{(a)} The one-axis protocol coefficient $\alpha_P$ on the 1{,}440 grid fits; all 12 runs share
the sign in every kind.
\textbf{(b)} Two-axis coefficients fitted per kind through the origin on the 2{,}016 two-axis fits. On
the axis that carries each kind's identity the coefficient is 0.75--0.87 ($q$, $k$, $o$, gate, up,
down); $v$, whose two fields are of similar size, has $\alpha_r = 0.42$ and $\alpha_c = 0.88$. The
minor-axis coefficient is poorly identified by these data: the minor-axis width has per-kind medians of
0.06--0.12, its square is 2--7 times below the major term, and the fitted minor coefficients include
negative values (gate $\alpha_c = -1.14$, down $\alpha_r = -1.13$); ranges beyond $\pm 2.5$ are truncated.
\textbf{(c)} Each model is fitted per kind. The column term is what brings $v$, $o$ and down into the
same relation; it changes $q/k$ little and slightly worsens gate/up, and the cross term changes the error by
at most 0.001 in every kind. The worst fold is a high-repetition arm in every kind and model, the shuffled one
(D3) in all but two cases.}
\label{fig:twoaxis}
\end{figure}

The balancing factors of eq.~\eqref{eq:decomp} give a consistency check on the additive form:
the variance of $\log|W_{ij}| = \log s + \log (D_r)_{ii} + \log (D_c)_{jj} + \log|Z_{ij}|$ splits into the two balancing-factor terms
and the core term with a cross-covariance share whose per-kind median is at most 0.3\% on the 2{,}016 grid
matrices (per-matrix maximum 3.3\%; per-matrix values in the data release, Appendix~\ref{app:codedata}). The bridge itself is fitted on
the direct marginal widths~\eqref{eq:fields}, which differ from these factors by the opposite-side term of
eq.~\eqref{eq:factorsfields}, and remains an empirical relation.

The bridge on both axes, eq.~\eqref{eq:bridge2}, is
an empirical through-origin approximation and not a theorem about Weibull mixtures. The
one-axis form~\eqref{eq:bridge}, whose baseline is $\krow$ rather than $\kbi$, is its one-axis counterpart
on the identity axis, which is the form the data identify. The second-axis term is not identifiable where the second field is small --- its sign agrees in
only 2--3 of 12 runs for $q/k$, gate and up --- and it is required where the second field is not small:
adding it lowers the leave-one-run-out error from 0.0060 to 0.0032 for $v$, from 0.0109 to 0.0042 for $o$
and from 0.0235 to 0.0071 for down, and changes $q/k$ little. A single pooled two-axis fit over all
seven kinds reaches only $R_0^2 = 0.90$ with unequal coefficients (0.67 rows, 0.46 columns), so the
one-axis bridge remains the main result and the two-axis form is its extension to the
column-identity kinds. Adding a cross term $\gamma\,\Hrow\Hcol$ as a robustness check changes the
leave-one-run-out error by at most 0.001 in every kind (Figure~\ref{fig:twoaxis}c).

On synthetic scale fields injected into real matrices the exact-Weibull coefficient $6/\pi^2$ predicts
the refitted shape $k$ better than 0.791 does, and on the grid the excess of $q/k$ over $6/\pi^2$ is carried by
the scale-times-within-row-shape pairing term; that is an association within the protocol, not a cause.

The width does not fix the departure: a scale-times-within-row-shape pairing term carries 12--28\% of
$\dkmix$ and varies with the data condition, and the local coefficient bends with $\HW$ in a way we report
and do not explain.

\subsubsection{Information retained beyond field width}

\begin{figure}[tbp]
\centering
\includegraphics[width=\textwidth]{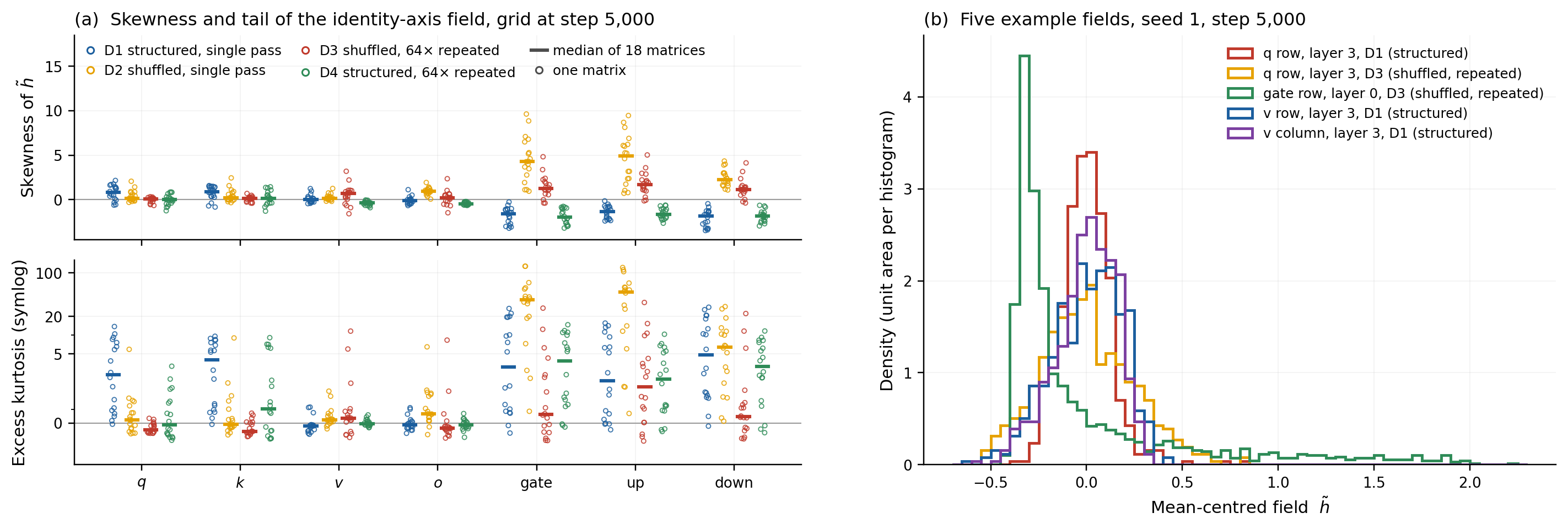}
\caption{\textbf{Scale-field distributions retain information beyond field width.} Controlled data grid
(Table~\ref{tab:configs}), 504 matrices at step 5{,}000, identity-axis field of each matrix. The plotted
quantity is $\tilde h$, the field $h$ of~\eqref{eq:fields} re-centred by its mean rather than by its median; the shift changes neither width, skewness nor
kurtosis, and $\tilde h = 0$ is the geometric-mean channel RMS.
\textbf{(a)} Positive skewness is a tail of amplified channels, negative a tail of suppressed channels.
\textbf{(b)} Channel histograms of five example matrices, chosen to show the range of shapes and not
matched in width.}
\label{fig:fielddist}
\end{figure}

Field width explains most of the pooled-shape departure but does not specify the field. Two synthetic
checks on the controlled grid (Table~\ref{tab:configs}) separate what it leaves out. Shape: four synthetic
field distributions of equal width (log-normal, two-point, sparse, arcsine) applied to the same $q/k$
cores give fitted shape values that spread increasingly with width, by 0.0015, 0.007 and 0.023 at
$\HW = 0.10$, 0.20 and 0.35, against a width-driven drop of 0.005, 0.022 and 0.068. Order: reordering a
fixed synthetic field over the channels changes the shape $k$ by at most 0.0002 in this test; that does not remove
the scale--core pairing term observed in real matrices above, which is a different quantity. The
quantitative evidence is therefore synthetic. Figure~\ref{fig:fielddist} adds the empirical fact on real
weights: the skewness and tails of the fields vary with the data condition, particularly in the FFN
projections: the FFN fields change the sign of their skewness and become heavy-tailed with the data arm,
the $q/k$ fields are heavy-tailed mainly in the structured arm D1, and the $v/o$ fields stay mostly
light-tailed.

\subsection{Functional axes and RoPE}
\label{app:axes}

This section gives the controls for the cross-projection field matching and the RoPE reassignment and
no-positional-encoding comparisons of Section~\ref{sec:identity}.

\subsubsection{When cross-projection matching persists or weakens}

Figure~\ref{fig:transfer} reports the grid-endpoint pairings and their null controls; this section adds
what the figure does not show. Before training, the pairings are absent: at step 0 every row-level
correlation across the three initialization families lies within 0.09 of zero. The longer
initialization-family runs show which matches persist. At 30{,}000 steps, up rows and down columns
correlate at about 0.90 in each family, and $v$ rows and $o$ columns at 0.95--0.96, whereas gate--down
and gate--up correlations fall to about 0.35 and 0.4, so the persistent FFN pairing is specifically
up--down. Within the 5{,}000-step data grid, matching is weakest in D2 (fully shuffled, single pass):
gate--down is 0.66 and $v/o$ is 0.26. The equally shuffled but highly repeated D3 arm reaches 0.95 and
0.91. The strength of a shared-channel match therefore changes with the data condition, and the D2
medians still exceed their within-layer re-pairing controls.

\subsubsection{RoPE assignment and removal: field organization versus formation}

\begin{figure}[tbp]
\centering
\includegraphics[width=0.9\textwidth]{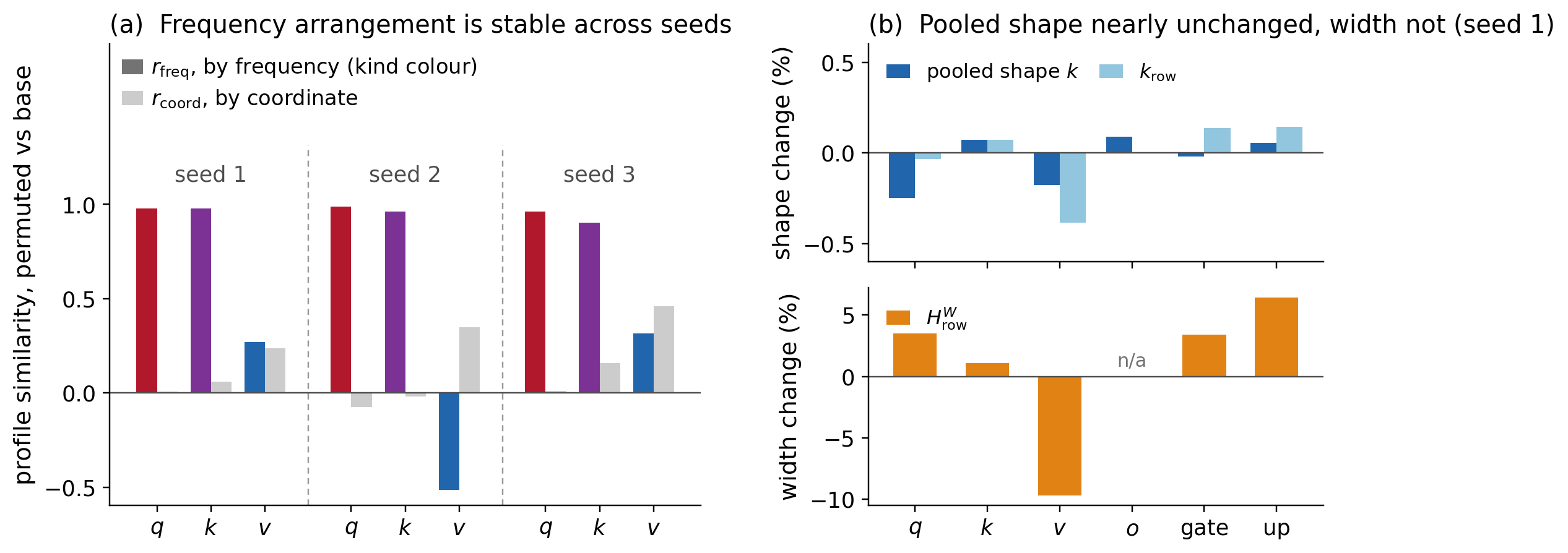}
\caption{\textbf{The RoPE permutation changes field arrangement while the pooled shape remains stable.}
Paired runs with base and reassigned RoPE frequencies, three seeds, same data and initialization.
\textbf{(a)} Similarity between the permuted and the base row profile at each paired seed, pooled over
heads and layers on the checkpoints shared by each pair (steps 3{,}200, 10{,}000 and 30{,}000); the pooled
$\Delta r_{\mathrm{id}}$ is the difference of the two bars. \textbf{(b)} Change under the permutation at
seed 1 (the width change differs across seeds, see text), each bar a pooled median over 18 matrices per
kind (6 layers $\times$ steps 20{,}000, 25{,}000 and 30{,}000); $o$ is not part of the row-side width
comparison because its identity axis is the column axis.}
\label{fig:rope_readout}
\end{figure}

Section~\ref{sec:identity_rope} establishes on one representative pair that the $q/k$ row profiles follow
the reassigned frequencies. Pair-level profiles average the log row RMS of the two rows of a rotary
pair (row $j$ and row $j$ plus half the head dimension) and take the median over heads. This average is not exactly invariant under the
common rotation of Section~\ref{sec:identity_rope}, whereas the log of the pair's pooled RMS is; on the
grid endpoints of Figure~\ref{fig:transfer}(c) the two give median $q/k$ correlations of 0.929 and 0.936
and gain-to-balance ratios of 22.2 and 22.0. Figure~\ref{fig:rope_readout} extends this to every paired seed and shared
checkpoint and separates what the reassignment leaves unchanged from what it moves. Reassigning the same
RoPE frequencies to different rows changes the coordinate-indexed $q/k$ profiles, while indexing by the
assigned frequencies restores their agreement across the three paired seeds; $v$ shows no comparable
separation (Figure~\ref{fig:rope_readout}a), although at seed 3 its pooled value ($-0.15$) lies above its one-sided
null ($-0.17$), so the pre-specified negative-control criterion is not met at that seed. Of the scalar read-outs, the pooled shape $k$ and $\krow$
change by less than 0.5\%, whereas the row width is not equally stable (Figure~\ref{fig:rope_readout}b):
its change depends on the statistic (sd 1--4\% against the interquartile read-out 7--10\%
for $q/k$), on the aggregation ($v$: $-9.7\%$ pooled over 18 layer--checkpoint matrices, $-5.1\%$ as the median
over checkpoints of layer medians, $-2.5\%$ at 30{,}000 steps alone) and
on the seed ($q$ at 30{,}000 steps: $+2\%$, $+26\%$, $+14\%$). The full field thus retains the
frequency-to-row assignment that the scalar read-outs largely omit. The held-out loss differs between the
two arms by 0.0013 nat.

\begin{figure}[tbp]
\centering
\includegraphics[width=\textwidth]{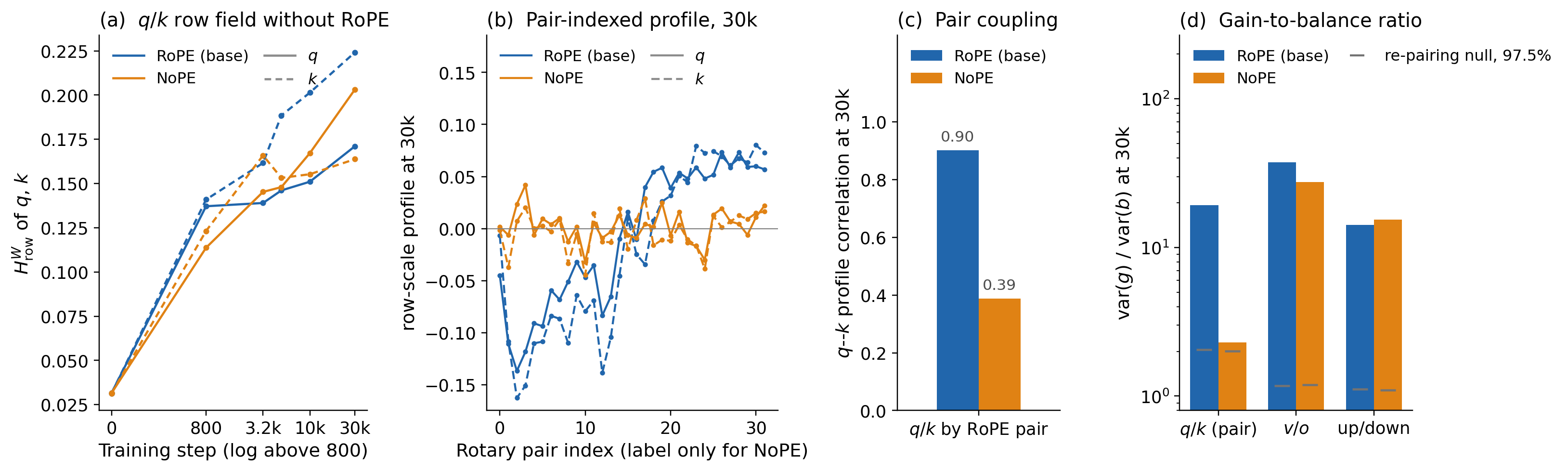}
\caption{\textbf{A $q/k$ row field forms without RoPE, but its rotary-pair organization and the
$q$--$k$ pair coupling weaken.} One run trained without positional rotation (rotary $\cos = 1$,
$\sin = 0$) against the base run: single seed, same data and initialization, terminal loss 3.174 against
2.977. In (d), $g = h_1 + h_2$ and $b = h_1 - h_2$ are the sum and difference of the two paired
profiles on each shared path (Section~\ref{sec:gainedit}), used here only as a pairing read-out; its grey
marks are the upper 97.5\% quantile of 100 within-layer re-pairings.}
\label{fig:nope}
\end{figure}

Removing RoPE gives a complementary comparison (Figure~\ref{fig:nope}). In this run, $q$ and $k$ row
fields still develop, but the pair-indexed $q$--$k$ coupling weakens from 0.90 to 0.39, above its re-pairing
null (97.5\% quantile 0.17 for the six-layer median; three of six layers exceed their per-layer null), and the
pair gain-to-balance ratio falls from 19 to 2.3, just above its null of 2.0; the row-level $q$--$k$
correlation rises from 0.26 to 0.85. The $v/o$ and up/down paths remain well above their nulls. RoPE therefore organizes the frequency-level $q/k$ relationship in these
runs, while formation of a row-scale field does not require that coordinate. The identity-axis widths of the
other kinds in this run at steps 3{,}200 and 30{,}000 ($v$ 0.185/0.186, $o$ 0.210/0.192, gate 0.216/0.097,
up 0.202/0.066, down 0.205/0.056) show the $v/o$ fields holding their level whereas the FFN fields recede
from a higher peak, as on the base run.

\paragraph{Boundaries.}
The reassignment keeps the frequency set fixed, whereas the run without positional encoding is one
higher-loss seed; neither comparison identifies the dynamics that generate the fields.

\subsection{Weight-field trajectories}
\label{app:dynamics}

\begin{figure}[tbp]
\centering
\includegraphics[width=\textwidth]{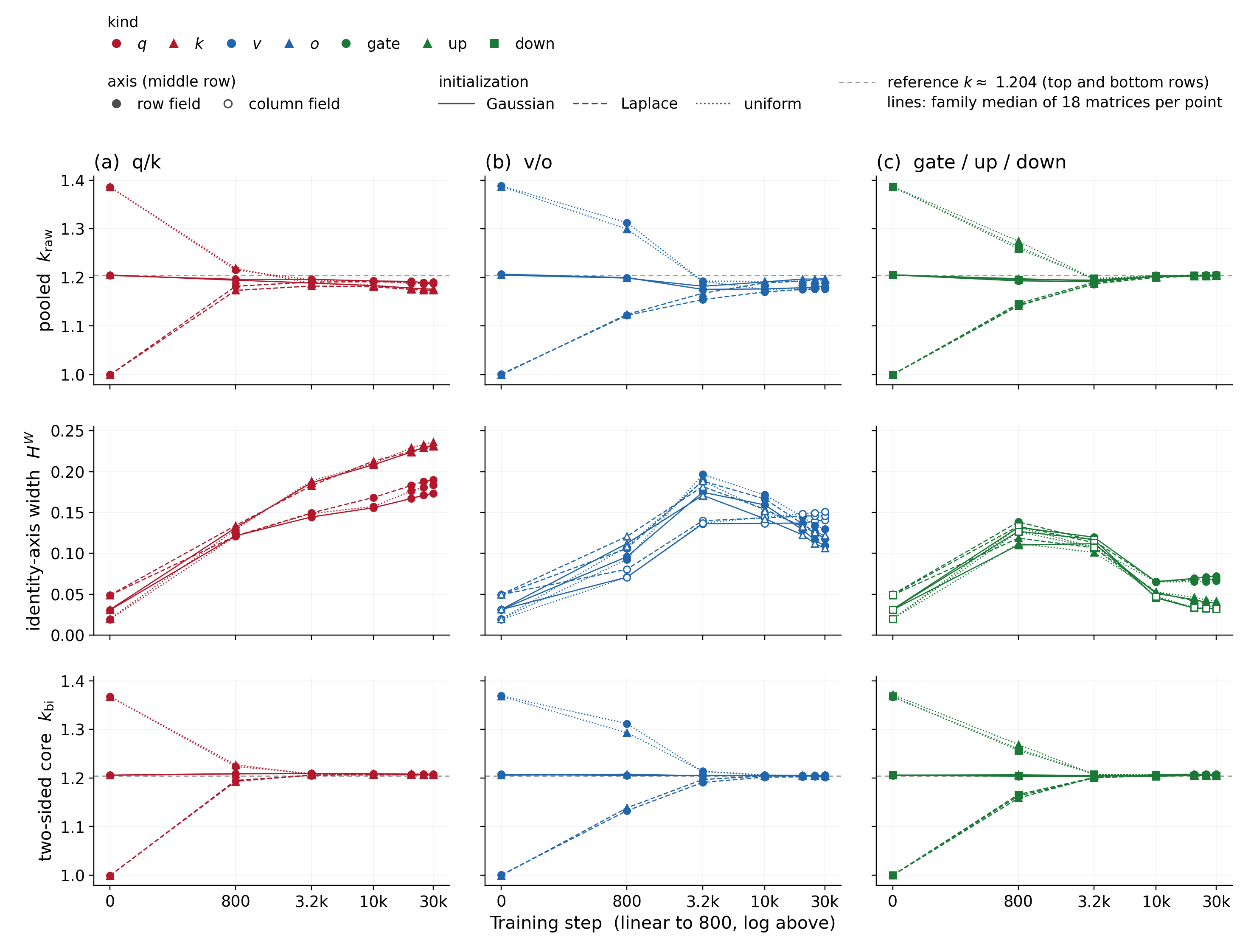}
\caption{\textbf{Core convergence and scale-field formation unfold together across initializations.}
Nine LLaMA-style 70M runs on one data setting (Gaussian, Laplace and uniform initialization $\times$ three
seeds), in three columns of kinds ($q/k$, $v/o$, gate/up/down). Top row: pooled shape $\kraw$; middle row:
identity-axis field width ($\Hrow$ for $q$, $k$, gate, up; $\Hcol$ for $o$, down; both for $v$); bottom row:
two-sided core shape $\kbi$. Lines are family medians over 18 matrices (three seeds $\times$ six layers).}
\label{fig:timing}
\end{figure}

This appendix checks the field trajectories of Section~\ref{sec:evolution} across Gaussian, Laplace and
uniform initializations. Figure~\ref{fig:evolution} follows three seeds of one family; Figure~\ref{fig:timing}
places the same nine runs of Figure~\ref{fig:core}(c) side by side in three read-outs, so that field formation
can be seen against core convergence in every family.

In the Gaussian runs of Figure~\ref{fig:evolution}, gate falls from its step-800 peak to 0.065 by step
10{,}000 and rebounds slightly to 0.072, while up and down narrow by $2.8\times$ and $3.9\times$ from their
peaks, down to within 4\% of its initialization width and up to 27\% above it. The bottom row of
Figure~\ref{fig:timing} gives the two-sided core; for $q/k$ it is 1.21 preserved under
Gaussian initialization, 1.00 and 1.37 at step 0 under Laplace and uniform, within 0.01 of the Gaussian
value by step 3{,}200, and 1.195--1.218 per matrix at 30{,}000 steps for all three families. Once the cores
have merged, the terminal pooled shape $k$ of each kind agrees to within 0.01 across the three initializations.
The middle row shows that the field is already well formed while the cores are still merging, and that the
formation--retention split of Figure~\ref{fig:evolution} (accumulating $q/k$, receding FFN, $v/o$ between)
appears in each family.

\subsection{AdamW second-moment factors}
\label{app:vfactors}

This appendix gives the update rule, the motivation and fits of Equation~\eqref{eq:vfactors}, and the
timing comparison behind Section~\ref{sec:optimizer}.

\paragraph{The AdamW update.}
For a projection $W_t \in \mathbb{R}^{d_{\mathrm{out}} \times d_{\mathrm{in}}}$ with mini-batch gradient $G_t$,
the AdamW state \citep{kingma2015adam, loshchilov2019decoupled} is
\begin{equation}
M_t = \beta_1 M_{t-1} + (1-\beta_1)\, G_t, \qquad
V_t = \beta_2 V_{t-1} + (1-\beta_2)\, G_t^{\odot 2},
\label{eq:adamw_state}
\end{equation}
with $M_0 = V_0 = 0$, and the weights advance as
\begin{equation}
\hat M_t = \frac{M_t}{1-\beta_1^t}, \qquad \hat V_t = \frac{V_t}{1-\beta_2^t}, \qquad
U_t = \frac{\hat M_t}{\sqrt{\hat V_t} + \epsilon}, \qquad
W_{t+1} = (1 - \eta_t \gamma_{\mathrm{wd}})\, W_t - \eta_t\, U_t,
\label{eq:adamw_update}
\end{equation}
where all operations are elementwise, $\epsilon$ is the stabilizer (distinct from the fit residual
$\varepsilon_{ij}$ below), $\eta_t$ the scheduled learning rate and $\gamma_{\mathrm{wd}}$ the decoupled weight decay;
the values used are in Appendix~\ref{app:configs}. Equation~\eqref{eq:vfactors} is fitted to the dumped $\hat V_t$; the bias correction multiplies
every coordinate by the same number and therefore shifts $\mu$ only.

\paragraph{Why an additive model in log space.}
For a linear projection $y = Wx$ the gradient contributed by a single token is the outer product
$\delta x^{\top}$, whose squared entries $\delta_i^2 x_j^2$ separate into a row and a column factor. The
mini-batch gradient is a sum of such outer products, $G_{ij} = \sum_u \delta_i^{(u)} x_j^{(u)}$, so
$G_{ij}^2$ also contains cross-token terms $\delta_i^{(u)}\delta_i^{(v)} x_j^{(u)} x_j^{(v)}$, and $\hat V$
further averages these squares over time. The per-token structure motivates row and column effects in the
accumulated second moment; if persistent row- and column-dependent parts dominate,
$\hat V_{ij} \approx C_V A_i B_j$, which is Equation~\eqref{eq:vfactors} after taking logarithms and centring.
The cross-token terms, the coupling of errors and activations, the attention mixing, the nonlinearities and
the coordinate-specific history remain in $\varepsilon_{ij}$, and the adequacy of the model is judged by its
$R^2$. The factors $a_i$ and $b_j$ are thus the output- and input-channel factors of the accumulated
squared-gradient state, indexed by channel, which is why they group by tensor (Figure~\ref{fig:topology}).
With one-sided $R^2$ values $d_{\mathrm{in}} \sum_i a_i^2 / \mathrm{SS}_{\mathrm{tot}}$ and
$d_{\mathrm{out}} \sum_j b_j^2 / \mathrm{SS}_{\mathrm{tot}}$ (per-kind medians over the 30 fits of six layers and
five dump steps), the full model reaches 0.86--0.98; the row factor alone explains 0.80 and 0.83 for $q$ and
$k$ and 0.68 and 0.69 for gate and up, the column factor alone 0.61 for $o$ and 0.85 for down, and $v$ splits
at 0.40 and 0.44, so the dominant factor is the identity axis of each kind.

\paragraph{The remainder shows weak coordinate-level association with the weights.}
The remainder $\varepsilon_{ij}$ holds 0.2--10\% of the variance of $\log \hat V$ at step 10{,}000 and becomes
heavy-tailed as training proceeds in every kind except $o$ (excess kurtosis below 1 at step 200; 7--22 at step
10{,}000, while $o$ stays near 1).
Double-centring $\log|W_{ij}|$ by its row and column means removes the same two axes from the weights, so the two
residuals can be compared coordinate by coordinate within one matrix. Their Spearman correlation, measured at the
same checkpoint, is weak: over the 210 matrix--step fits of the five-dump run of Figure~\ref{fig:topology}
the median $|\rho|$ is 0.017 and the maximum
0.089, against a coordinate-permutation null below 0.006; 21 fits exceed 0.05, all negative, on gate, $k$, $v$ and
$o$ at intermediate steps. In this contemporaneous comparison the correspondence between optimizer state
and weight fields is carried by the row and column axes; a lagged influence through the update is not
tested.

\paragraph{What the factors mean for the update.}
Ignoring the stabilizer, the AdamW step is $U_{ij} = \hat M_{ij}/\sqrt{\hat V_{ij}}$, so under
Equation~\eqref{eq:vfactors} the denominator factorizes as $e^{\mu/2} e^{a_i/2} e^{b_j/2}$ up to
$e^{\varepsilon_{ij}/2}$: for a fixed first moment, a positive $a_i$ damps every update in row $i$ and a
positive $b_j$ every update in column $j$. The realized update also depends on $\hat M_{ij}$, so the factors describe the geometry of the
denominator, not the complete step. Adafactor \citep{shazeer2018adafactor} shares the row--column viewpoint but reconstructs $\hat V$
from original-space marginals to save memory; Equation~\eqref{eq:vfactors} is a descriptive fit in log space,
and a high log-space $R^2$ does not make $\hat V$ rank one.

\paragraph{Cells of Figure~\ref{fig:topology}.}
The gate and up columns, which read the same FFN input, correlate at 0.998 from the first dump. The
attention-input and FFN-input columns, attached to the same residual coordinate, correlate at 0.53--0.76 at
step 800 and at 0.00--0.07 by step 10{,}000. The FFN-input columns and the rows carrying the residual-output
error rise from 0.26--0.29 at step 200 to 0.45--0.54 at step 10{,}000, whereas the attention-input columns
stay within $\pm 0.10$ of both. The permutation null of 0.07 is a median over cells of cell-specific 95th
percentiles, not a single threshold for the whole matrix.

The dump-based factors above are not interchangeable with the logger profile
$\log\sqrt{\mathrm{mean}\,\hat V}$ along an axis, used for the time-resolved analyses, which differs most for
the row-level $q/k$ pair; the timing below refers to logger profiles. They come from the separate logger run of
Table~\ref{tab:configs} (batch size 16, its own window sequence), which records axis-wise $W$, $\hat M$ and
$\hat V$ statistics over a 20-step window every 200 steps to 30{,}000 steps. Taking as lock-in the first step at
which a row profile reaches a Spearman correlation of 0.7 with its profile in the final window (steps
29{,}801--29{,}820), the $\hat V$ row profile locks in before the weight row field on six of seven kinds, by
0.7k--1.5k steps on $q$, $v$ and $o$ and by 3.7k--5.9k on $k$, up and down. Gate is the exception:
its weight field locks in at 4.8k steps while its $\hat V$ profile locks in at 9.5k. The reversal is not
specific to $\hat V$: every gradient-side quantity feeding gate (the error rows, the gradient, $\hat M$ and
$\hat V$) locks in between 9.5k and 16.7k steps, so the gradient statistics reaching gate keep reorganizing
long after gate's own row profile has settled. A candidate reading is the SwiGLU product \citep{shazeer2020glu}: the gradient
reaching a gate row is weighted by the paired up activation and inherits the up field, which recedes until
about 10k steps, whereas up's gradient inherits the early-settling gate field and locks in at 5k--6k steps,
close to gate's weight lock-in. Cross-matrix pairing runs
the other way: $v$ rows and $o$ columns correlate above 0.7 at 676 steps in $W$ but only at 3{,}528 steps
in $\hat V$. Neither state therefore uniformly precedes the other, and the gate reading is a single-run
candidate.

\subsection{Paired gain and frozen-checkpoint edits}
\label{app:edits}
\label{app:gainedit}

This appendix gives the gain--balance decomposition, the edit protocol and the controls behind
Section~\ref{sec:gainedit}.

\begin{figure}[tbp]
\centering
\includegraphics[width=\textwidth]{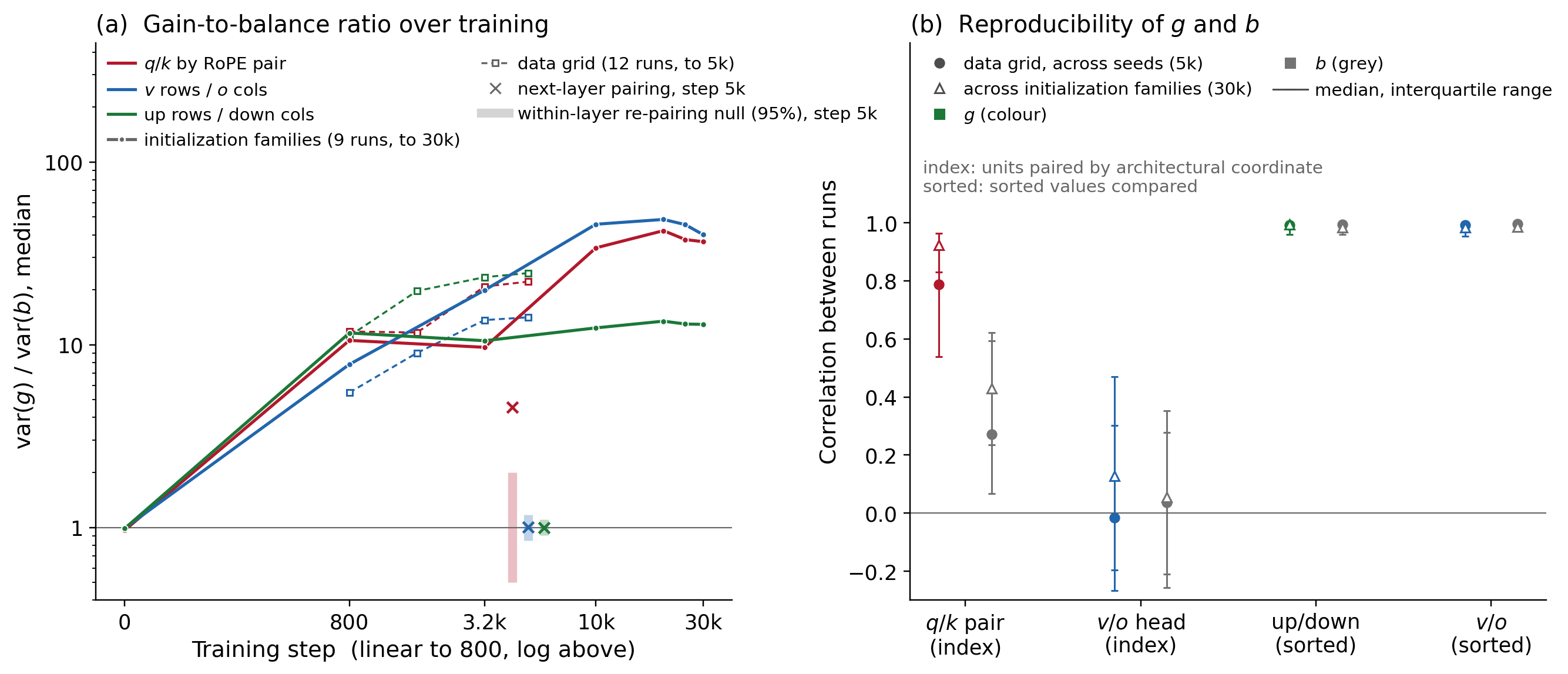}
\caption{\textbf{Matched scale fields accumulate predominantly in a shared gain mode.} Gain $g$ and
balance $b$ of a paired channel are defined in Section~\ref{sec:gainedit}; their variance ratio exceeds 1
when the two profiles move together. Re-paired profiles give about 1 (0.5--2.0 on $q/k$); the adjacent-layer
$q/k$ pairing stays near 4.6. Both run sources are defined in
Table~\ref{tab:configs}. \textbf{(a)} The re-pairing null (shaded bars) and the next-layer pairings (crosses) are computed at step
5{,}000 of the data grid only and are drawn there, one per path.
\textbf{(b)} Head order is a permutation-symmetric label, so $v/o$ has no reproducible index; up/down is
shown as a sorted comparison only.}
\label{fig:gain}
\end{figure}

\paragraph{Paired gain and balance modes.}
For two projections sharing a unit identity, with centred log-scale profiles $h_1, h_2$, each unit has a gain
mode $g = h_1 + h_2$ and a balance mode $b = h_1 - h_2$. Since
$\mathrm{var}(g)-\mathrm{var}(b)=4\,\mathrm{cov}(h_1,h_2)$, their variance ratio reads the cross-channel
matching; the edits below test function. At 5{,}000 steps the ratio
$\mathrm{var}(g)/\mathrm{var}(b)$ is 22 for $q/k$ by RoPE pair, 14 for $v$ rows against $o$ columns and 25
for up rows against down columns, against 0.97--0.99 at step 0. Within-layer re-pairing gives 95\% null bands of
0.85--1.18 on $v/o$ and 0.91--1.11 on up/down, and pairing with the adjacent layer gives 1.00 on both; on $q/k$
the re-pairing band is wider (0.50--2.01) and the adjacent-layer pairing stays at 4.6, consistent with a
frequency profile shared across layers. The two sides of a matched channel move together
(Figure~\ref{fig:gain}a). On the initialization-family runs at 30{,}000 steps the gate/down and gate/up ratios fall to 1.7--2.3 while up/down
stays at 12--14, so the persistent FFN matching is on up/down, and the edits leave gate fixed. Where the
unit has an architectural coordinate the gain mode also aligns across runs: for $q/k$ by
RoPE pair its correlation is 0.79 across seeds and 0.92 across initialization families, against 0.27 and
0.43 for the balance mode (Figure~\ref{fig:gain}b); the $v/o$ head index does not align, whereas the sorted
profiles of both modes correlate at 0.97--1.00 per channel on every path (0.93--0.94 at the head level): the
shape of the gain profile is shared across runs, the assignment to heads is not. For the
$q/k$ and $v/o$ pairs the coupling echoes the balancing condition derived for regularized bilinear factors
\citep{kobayashi2024lowrank}; the gated FFN is not a bilinear
factorization, and its up/down matching is an empirical observation.

\paragraph{What is edited.}
The edits of Section~\ref{sec:gainedit} act on the final checkpoint of the base run of the paired RoPE reassignment (LLaMA-style 70M,
Gaussian initialization, seed 1, 30{,}000 steps). No parameter is trained; every condition is a
rewrite of the stored weights followed by forward evaluation, with fixed random seeds for the randomized controls. A path is a pair of projections
that share a channel: $v$ rows with $o$ columns (512 head/value channels per layer), up rows with down
columns (1{,}376 FFN hidden units per layer), and $q$ rows with $k$ rows taken at the RoPE-pair level
(rows $d$ and $d+32$ of each head form one channel with scale $\sqrt{(R_d^2 + R_{d+32}^2)/2}$, the
rotation-invariant pooled RMS; 256 pairs per layer). For each layer and channel $i$ the two identity-axis
fields $h_1(i)$ and $h_2(i)$ are the median-centred log scales of the two projections on the shared side, as
in eq.~\eqref{eq:fields}; $g$ and $b$ are not re-centred, and flattening sets the selected channels to the
median level $g = 0$ (the median of $g$ lies within 0.006 of zero on every path and layer). The gain is $g = h_1 + h_2$ and the balance is $b = h_1 - h_2$. An edit
multiplies the row (or column) $i$ of side 1 by $\phi_1(i)$ and of side 2 by $\phi_2(i)$:
\begin{center}
\begin{tabular}{@{}lll@{}}
\toprule
edit & $\phi_1(i)$ & $\phi_2(i)$ \\
\midrule
balance & $e^{-b(i)/2}$ & $e^{+b(i)/2}$ \\
flatten at strength $\tau$ & $e^{-\tau\,g(i)/2}$ & $e^{-\tau\,g(i)/2}$ \\
shuffle by permutation $\pi$ & $e^{(g(\pi(i))-g(i))/2}$ & $e^{(g(\pi(i))-g(i))/2}$ \\
shuffle, displacement-matched & $e^{(g(\pi(i))-g(i))/(2\sqrt 2)}$ & same \\
decile $m$ & $e^{-g(i)/2}$ if $g(i)$ in decile $m$, else 1 & same \\
Gaussian direction & $e^{-\xi(i)/2}$, $\xi \sim \mathcal N(0, \mathrm{var}(g))$ i.i.d. & same \\
random signs & $e^{-\sigma(i)\,g(i)/2}$, $\sigma(i) = \pm 1$ i.i.d. & same \\
\bottomrule
\end{tabular}
\end{center}
On the $q/k$ path $\phi$ is applied to both rows of a pair, so the path is edited at pair level and a
within-pair asymmetry is never touched; the gate projection is not edited. Every edit acts on all six layers
of one path at once and leaves the other five kinds untouched. After every gain edit each matrix is rescaled
to its original Frobenius norm, which holds the global scale $s$ fixed; positive diagonal scalings preserve
the balanced core $Z$, while the direct marginals on either axis can change
(Appendix~\ref{app:conversions}). Balance edits are not renormalized, since their invariance rests on the
exact reciprocal factors. At strength $\tau$ the gain becomes $(1-\tau)g$ up to channel-independent
constants: $\tau=1$ flattens it, $\tau=-1$ doubles it and $\tau=2$ reverses it. The size of an edit is its displacement, the relative squared Frobenius change summed over the twelve
matrices $W_m$ of the path, $\dF = \sum_m \|\Delta W_m\|_F^2 / \|W_m\|_F^2$; $\dF^{\mathrm{flat}}$ is that of
the flatten at $\tau = 1$. The loss read-out of an edit is $\Delta L$, the change in mean next-token loss
(nats per token) relative to the unedited model on the same tokens.

\paragraph{Why the balance edit is an identity.}
On the $v/o$ path, the attention-weighted value in each head is linear in its value channels.
Scaling row $i$ of $W_v$ by $c$ scales that channel at every token; the attention weights are unchanged
because $q/k$ are not edited. Scaling the corresponding column of $W_o$ by $1/c$ then cancels the change. The same
holds for up/down through the FFN hidden unit, except that the nonlinearity sits between the two sides:
SwiGLU \citep{shazeer2020glu} multiplies the up activation by the gated activation, so scaling up by $c$ and down by $1/c$ is
exact only because the gate is not edited and the activation is linear in the up side. For $q/k$ the score
$q^\top k$ of a head is a sum over RoPE pairs; rotation by the pair's frequency commutes with a scalar
applied to both rows of the pair, so scaling $q$'s pair by $c$ and $k$'s pair by $1/c$ leaves every score
unchanged. Biases are absent in this architecture. The balance edits change the loss at
the $10^{-7}$ level on every path and slice (at most $1.2\times10^{-7}$), the float32 resolution of the mean loss, which verifies the
pairing, the pair-level treatment of RoPE and the implementation for the forward pass.

\paragraph{Evaluation.}
The loss is the mean next-token cross-entropy over 48 sequences of 512 tokens, in batches of 8, read from
the pre-tokenized stream of $10^8$ tokens that the run was trained on. Slice 1 starts at offset
$9\times10^7$ and slice 2 at offset $6\times10^7$, without overlap; Figure~\ref{fig:gainedit} shows both,
and the absolute costs quoted in Section~\ref{sec:gainedit} are slice-1 values. Training
sampled the stream with replacement, so both slices are in-distribution; each edit is compared with the
unedited model on the same tokens.

\paragraph{Controls and reproducibility across slices.}
The flatten at $\tau=1$ is compared with four controls: the gain permutation scaled by $1/\sqrt{2}$ (at
its own width it displaces about twice as much as the flatten), a Gaussian direction in the same channel
subspace with the variance of $g$, the flatten with an independent random sign per channel, and the gain
doubled ($\tau=-1$). The controls lie at displacement ratios 0.87--1.52 of the flatten, the scaled
permutation at 0.98--1.00 (Table~\ref{tab:gainedit_matched}). Figure~\ref{fig:gainedit}(b) reports
the displacement-normalized response $(\Delta L/\Delta L^{\mathrm{flat}})/(\dF/\dF^{\mathrm{flat}})$, a
descriptive quantity for finite edits; the table gives its two factors separately.

After this normalization, flattening costs more than the Gaussian direction on both slices, by 3.2--3.3 on
$q/k$, 1.4--1.6 on $v/o$ and 2.0--3.1 on up/down. The scaled permutation is its $1/\sqrt 2$ projection onto
$-g$ plus a remainder of half the variance of $g$; to second order its cost relative to the flatten is
therefore $\tfrac12 + \tfrac12 \times$ the raw Gaussian cost ratio, which predicts 0.65, 0.81 and 0.75 on slice 1
and 0.64, 0.77 and 0.67 on slice 2, against the observed 0.64/0.66, 0.87/0.85 and 0.65/0.69. The flatten costs
themselves depart from a quadratic dependence on $\tau$, most on $q/k$, so the agreement is approximate; the
control gives no evidence of sensitivity to the channel assignment beyond that projection.

Across the two slices the balance identity, the positive flatten costs, the advantage over the Gaussian
direction and the endpoint decile with the larger response all reproduce. Absolute costs are higher on
slice 2, and the $q/k$ doubling asymmetry reproduces (3.64 and 3.48), whereas the FFN doubling response
is slice-dependent (1.59 and 0.64). The ten decile edits of $v/o$ sum to 0.68--0.70 of the full flatten,
so these finite responses are not additive. The four middle deciles change the loss by at most $6\times10^{-4}$ nats, against $2\times10^{-3}$ to 0.14 for the
two extreme deciles, and the
ratio of flatten cost to squared gain width moves with the slice and is kept only as a candidate.

\paragraph{Localization.}
Part of the decile concentration of Section~\ref{sec:gainedit} follows from the larger displacement at the
extremes; after displacement normalization a concentration of 1.6 remains at the amplified end of $q/k$ and
1.2--1.4 at the suppressed end of up/down, while the amplified end of $v/o$ falls to 0.8, below its share of
the displacement. Of the predictions recorded before the edit runs, the identity and the monotone cost hold, the
permutation's arrangement reading was withdrawn after the displacement comparisons, and the amplified-end
prediction reverses on the FFN path.

\begin{table}[H]
\centering\footnotesize
\setlength{\tabcolsep}{3.5pt}
\begin{tabular}{@{}l rrr rrr rrr@{}}
\toprule
& \multicolumn{3}{c}{$q/k$} & \multicolumn{3}{c}{$v/o$} & \multicolumn{3}{c}{up/down} \\
\cmidrule(lr){2-4}\cmidrule(lr){5-7}\cmidrule(lr){8-10}
edit & size & slice 1 & slice 2 & size & slice 1 & slice 2 & size & slice 1 & slice 2 \\
\midrule
flatten (reference) & 1.00 & 1.00 & 1.00 & 1.00 & 1.00 & 1.00 & 1.00 & 1.00 & 1.00 \\
\midrule
log-gain doubled ($\tau=-1$) & 1.23 & 3.64 & 3.48 & 1.20 & 1.48 & 1.16 & 0.99 & 1.59 & 0.64 \\
log-gain $\times1.5$ ($\tau=-0.5$) & 0.29 & 1.30 & 1.31 & 0.30 & 0.38 & 0.28 & 0.25 & 0.45 & 0.10 \\
shuffle, size-matched (5 draws) & 0.98 & 0.64 & 0.66 & 1.00 & 0.87 & 0.85 & 1.00 & 0.65 & 0.69 \\
Gaussian direction (5 draws) & 0.94 & 0.29 & 0.29 & 0.87 & 0.62 & 0.54 & 1.01 & 0.50 & 0.33 \\
random signs (5 draws) & 1.23 & 1.47 & 1.43 & 1.52 & 1.01 & 0.81 & 0.99 & 1.03 & 0.78 \\
\midrule
shuffle at the histogram's width (5 draws) & 1.96 & 1.13 & 1.14 & 1.98 & 1.82 & 1.76 & 1.99 & 1.45 & 1.31 \\
\midrule
\multicolumn{10}{@{}l}{\textbf{Read-outs of the 63-condition set (cost relative to the flatten unless stated)}} \\
\midrule
flatten cost, nats & 1.00 & 0.143 & 0.164 & 1.00 & 0.071 & 0.084 & 1.00 & 0.010 & 0.016 \\
gain reversed ($\tau=2$) & 3.71 & 2.73 & 2.78 & 3.40 & 3.32 & 3.43 & 4.05 & 5.35 & 4.14 \\
most amplified decile alone & 0.50 & 0.79 & 0.82 & 0.63 & 0.48 & 0.48 & 0.28 & 0.24 & 0.28 \\
most suppressed decile alone & 0.22 & 0.13 & 0.11 & 0.13 & 0.06 & 0.08 & 0.38 & 0.47 & 0.53 \\
sum of the ten deciles & -- & 0.96 & 0.99 & -- & 0.68 & 0.70 & -- & 0.82 & 0.90 \\
flatten cost / squared gain width & -- & 1.9 & 2.2 & -- & 1.6 & 1.9 & -- & 2.2 & 3.4 \\
\bottomrule
\end{tabular}
\caption{\textbf{Control-edit displacements and loss responses.} Unless marked as an absolute quantity, the slice columns report the change in loss of the edit relative to flattening the gain ($\tau=1$) on the same path and slice, averaged over the five random draws where marked; size is the squared displacement ratio $\dF/\dF^{\mathrm{flat}}$ of Appendix~\ref{app:gainedit}. The shuffle at the histogram's width is the unscaled permutation of the edit table in Appendix~\ref{app:gainedit}. The lower block reads the 63-condition set of the same two slices: the flatten cost in nats (base loss 3.347 on slice 1, 3.093 on slice 2), and, relative to it, the reversal $\tau=2$, the two extreme deciles of $g$ edited alone, the sum of all ten decile edits, and the flatten cost divided by the squared gain width (0.275, 0.209, 0.068 on the three paths).}
\label{tab:gainedit_matched}
\end{table}

\end{document}